\documentclass{ar-1col-S2O}
\usepackage{url}
\usepackage[style=ieee,backend=biber,sorting=none,maxnames=6,minnames=3,maxcitenames=1,mincitenames=1,url=false,doi=false,autolang=other]{biblatex}

\usepackage{siunitx}
\usepackage{tikz}
\usepackage{pgfplots}
\pgfplotsset{compat=newest}
\usepgfplotslibrary{groupplots}
\usepackage{graphicx}
\usepackage{hyperref}

\usepackage{cleveref}
\usepackage{aircraftshapes}
\DeclareSIUnit\foot{ft}

\usetikzlibrary{3d,arrows.meta,decorations.pathreplacing,calc}
\makeatletter
\tikzoption{canvas is xy plane at z}[]{%
  \def\tikz@plane@origin{\pgfpointxyz{0}{0}{#1}}%
  \def\tikz@plane@x{\pgfpointxyz{1}{0}{#1}}%
  \def\tikz@plane@y{\pgfpointxyz{0}{1}{#1}}%
  \tikz@canvas@is@plane
}
\makeatother

\makeatletter
\let\OldIncludeGraphics\includegraphics
\renewcommand{\includegraphics}[2][]{%
  \begingroup
    \let\centerline\@firstofone
    \let\leftline\@firstofone
    \OldIncludeGraphics[#1]{#2}%
  \endgroup
}
\makeatother

\newcommand{\casgrid}{%
	\fill[black!10, rounded corners=4, draw=black] (-0.6,-0.6) rectangle ++ (4.2,5.2);
	\fill[white, rounded corners=2, draw=black!50] ($(0,0) - (0.35,0.4)$) rectangle ++ (0.7,0.8);
	\fill[white, rounded corners=2, draw=black!50] ($(0,1) - (0.35,0.4)$) rectangle ++ (0.7,0.8);
	\fill[white, rounded corners=2, draw=black!50] ($(0,2) - (0.35,0.4)$) rectangle ++ (0.7,0.8);
	\fill[white, rounded corners=2, draw=black!50] ($(0,3) - (0.35,0.4)$) rectangle ++ (0.7,0.8);
	\fill[white, rounded corners=2, draw=black!50] ($(0,4) - (0.35,0.4)$) rectangle ++ (0.7,0.8);
	\fill[white, rounded corners=2, draw=black!50] ($(1,0) - (0.35,0.4)$) rectangle ++ (0.7,0.8);
	\fill[white, rounded corners=2, draw=black!50] ($(1,1) - (0.35,0.4)$) rectangle ++ (0.7,0.8);
	\fill[white, rounded corners=2, draw=black!50] ($(1,2) - (0.35,0.4)$) rectangle ++ (0.7,0.8);
	\fill[white, rounded corners=2, draw=black!50] ($(1,3) - (0.35,0.4)$) rectangle ++ (0.7,0.8);
	\fill[white, rounded corners=2, draw=black!50] ($(1,4) - (0.35,0.4)$) rectangle ++ (0.7,0.8);
	\fill[white, rounded corners=2, draw=black!50] ($(2,0) - (0.35,0.4)$) rectangle ++ (0.7,0.8);
	\fill[white, rounded corners=2, draw=black!50] ($(2,1) - (0.35,0.4)$) rectangle ++ (0.7,0.8);
	\fill[white, rounded corners=2, draw=black!50] ($(2,2) - (0.35,0.4)$) rectangle ++ (0.7,0.8);
	\fill[white, rounded corners=2, draw=black!50] ($(2,3) - (0.35,0.4)$) rectangle ++ (0.7,0.8);
	\fill[white, rounded corners=2, draw=black!50] ($(2,4) - (0.35,0.4)$) rectangle ++ (0.7,0.8);
	\fill[white, rounded corners=2, draw=black!50] ($(3,0) - (0.35,0.4)$) rectangle ++ (0.7,0.8);
	\fill[white, rounded corners=2, draw=black!50] ($(3,1) - (0.35,0.4)$) rectangle ++ (0.7,0.8);
	\fill[white, rounded corners=2, draw=black!50] ($(3,2) - (0.35,0.4)$) rectangle ++ (0.7,0.8);
	\fill[white, rounded corners=2, draw=black!50] ($(3,3) - (0.35,0.4)$) rectangle ++ (0.7,0.8);
	\fill[white, rounded corners=2, draw=black!50] ($(3,4) - (0.35,0.4)$) rectangle ++ (0.7,0.8);
}

\newcommand{\colorit}[2]{%
	\fill[#1, rounded corners=2, draw=black!50] ($(#2) - (0.35,0.4)$) rectangle ++ (0.7,0.8);
}

\newcommand{\coloritopacb}[4]{%
	\fill[#1, rounded corners=2, draw=black, fill opacity=#3, line width=#4] ($(#2) - (0.35,0.4)$) rectangle ++ (0.7,0.8);
}

\newcommand{\labelclimb}[2]{%
    \fill[#1, rounded corners=2, draw=black!50] ($(#2) - (0.35,0.4)$) rectangle ++ (0.7,0.8);
    \draw[-stealth, line width=1pt] ($(#2) - (0.25,0.25)$) -- ($(#2) + (0.25,0.25)$);
}

\newcommand{\labeldescend}[2]{%
    \fill[#1, rounded corners=2, draw=black!50] ($(#2) - (0.35,0.4)$) rectangle ++ (0.7,0.8);
    \draw[-stealth, line width=1pt] ($(#2) - (0.25,-0.25)$) -- ($(#2) + (0.25,-0.25)$);
}

\newcommand{\threesplit}[1]{%
	\fill[white, rounded corners=2, draw=black!50] ($(#1) - (0.35,0.4)$) rectangle ++ (0.7,0.8);
	\draw[black!50] ($(#1) - (0.35,0.13333)$) -- ($(#1) + (0.35,-0.13333)$);
	\draw[black!50] ($(#1) - (0.35,-0.13333)$) -- ($(#1) + (0.35,0.13333)$);
	\node[font=\tiny] at ($(#1) + (0,0.26667)$) {climb};
	\node[font=\tiny] at ($(#1) + (0,0)$) {none};
	\node[font=\tiny] at ($(#1) + (0,-0.26667)$) {desc.};
}

\newcommand{\threesplitcolor}[4]{%
    \fill[#2, rounded corners=2, draw=black!50] ($(#1) - (0.35,0.4)$) rectangle ++ (0.7,0.8);
    \fill[#4, rounded corners=2, draw=black!50] ($(#1) - (0.35,0.4)$) rectangle ++ (0.7,0.5);
    \fill[#3, draw=black!50] ($(#1) - (0.35,0.13333)$) rectangle ++ (0.7,0.26667);

	\draw[black!50] ($(#1) - (0.35,0.13333)$) -- ($(#1) + (0.35,-0.13333)$);
	\draw[black!50] ($(#1) - (0.35,-0.13333)$) -- ($(#1) + (0.35,0.13333)$);
	\node[font=\tiny] at ($(#1) + (0,0.26667)$) {climb};
	\node[font=\tiny] at ($(#1) + (0,0)$) {none};
	\node[font=\tiny] at ($(#1) + (0,-0.26667)$) {desc.};
}

\newcommand{\threesplitcolorh}[4]{%
    \fill[#2, rounded corners=2, draw=black!50] ($(#1) - (0.35,0.4)$) rectangle ++ (0.7,0.8);
    \fill[#4, rounded corners=2, draw=black!50] ($(#1) - (0.35,0.4)$) rectangle ++ (0.7,0.5);
    \fill[#3, draw=black!50] ($(#1) - (0.35,0.13333)$) rectangle ++ (0.7,0.26667);

	\draw[black!50] ($(#1) - (0.35,0.13333)$) -- ($(#1) + (0.35,-0.13333)$);
	\draw[black!50] ($(#1) - (0.35,-0.13333)$) -- ($(#1) + (0.35,0.13333)$);
	\node[font=\tiny] at ($(#1) + (0,0.26667)$) {left};
	\node[font=\tiny] at ($(#1) + (0,0)$) {none};
	\node[font=\tiny] at ($(#1) + (0,-0.26667)$) {right};
}

\newcommand{\threesplitcolorbig}[4]{%
    \fill[#2, rounded corners=2, draw=black, line width=1.5pt] ($(#1) - (0.5425,0.62)$) rectangle ++ (1.085,1.24);
    \fill[#4, rounded corners=2, draw=black, line width=1.5pt] ($(#1) - (0.5425,0.62)$) rectangle ++ (1.085,0.775);
    \fill[#3, draw=black, line width=1.5pt] ($(#1) - (0.5425,0.2066615)$) rectangle ++ (1.085,0.4133385);

	\draw[black!50] ($(#1) - (0.5425,0.2066615)$) -- ($(#1) + (0.5425,-0.2066615)$);
	\draw[black!50] ($(#1) - (0.5425,-0.2066615)$) -- ($(#1) + (0.5425,0.2066615)$);
	\node[font=\footnotesize] at ($(#1) + (0,0.4133385)$) {climb};
	\node[font=\footnotesize] at ($(#1) + (0,0)$) {none};
	\node[font=\footnotesize] at ($(#1) + (0,-0.4133385)$) {desc.};
}

\definecolor{pastelMagenta}{HTML}{FF48CF} % magenta
\definecolor{pastelPurple}{HTML}{8770FE} % purple
\definecolor{pastelBlue}{RGB}{0,114,178} % blue 0072B2
\definecolor{pastelSkyBlue}{RGB}{86,180,233} % sky blue 56B4E9
\definecolor{pastelGreen}{RGB}{0,158,115} % green 009E73
\definecolor{pastelOrange}{RGB}{230,159,0} % orange E69F00
\definecolor{pastelRed}{HTML}{F5615C} % red
\definecolor{darkColor}{HTML}{300A24} % dark
\definecolor{pastelGreenBlue}{RGB}{43, 169, 174} % mix of pastelGreen and pastelSkyBlue
\definecolor{fignumcolor}{cmyk}{1.0,0.4,0,0}

\definecolor{ra_1}{rgb}{1.0, 1.0, 1.0}
\definecolor{ra_2}{rgb}{0.6, 0.8, 1.0}
\definecolor{ra_3}{rgb}{1.0, 0.7058823529411765, 0.7058823529411765}
\definecolor{ra_4}{rgb}{1.0, 0.3568627450980392, 0.3568627450980392}
\definecolor{ra_5}{rgb}{0.7450980392156863, 0.0, 0.0}

\jname{Xxxx. Xxx. Xxx. Xxx.}
\jvol{AA}
\jyear{2025}
\begin{document}

% Page header
\markboth{Katz et al.}{Aircraft Collision Avoidance Systems}

% Title
\title{Aircraft Collision Avoidance Systems: Technological Challenges and Solutions on the Path to Regulatory Acceptance}

%Authors, affiliations address.
\author{Sydney M. Katz,$^1$ Robert J. Moss,$^1$ Dylan M. Asmar,$^2$ Wesley A. Olson,$^2$ James K. Kuchar,$^2$ and Mykel J. Kochenderfer$^1$
\affil{$^1$Department of Aeronautics and Astronautics, Stanford University, Stanford, CA, USA, 94305; email: \{smkatz,mossr,mykel\}@stanford.edu}
\affil{$^2$Lincoln Laboratory, Massachusetts Institute of Technology, Lexington, MA, USA, 02421; email:\{dylan.asmar,kuchar,wes.olson\}@ll.mit.edu}}

%Abstract
\begin{abstract}
% Abstract text, approximately 150 words. 
Aircraft collision avoidance systems is critical to modern aviation. These systems are designed to predict potential collisions between aircraft and recommend appropriate avoidance actions. Creating effective collision avoidance systems requires solutions to a variety of technical challenges related to surveillance, decision making, and validation. These challenges have sparked significant research and development efforts over the past several decades that have resulted in a variety of proposed solutions. This article provides an overview of these challenges and solutions with an emphasis on those that have been put through a rigorous validation process and accepted by regulatory bodies. The challenges posed by the collision avoidance problem are often present in other domains, and aircraft collision avoidance systems can serve as case studies that provide valuable insights for a wide range of safety-critical systems.
\end{abstract}

%Keywords, etc.
\begin{keywords}
% keywords, separated by comma, no full stop, lowercase
aircraft collision avoidance, aviation safety, decision making, validation, regulation
\end{keywords}
\maketitle

%Table of Contents
% \tableofcontents

% Heading 1
\section{INTRODUCTION}

Aircraft collision avoidance systems play a vital role in maintaining the safety of our airspace. These systems are designed to predict potential collisions between aircraft and take appropriate actions to avoid them. As part of the broader airspace conflict management strategy, they represent the final layer of protection against mid-air collisions and serve as an independent safeguard to mitigate the risks of potential errors and gaps in the aviation system. This article provides an overview of the technological challenges and solutions in the development of these systems with a particular focus on solutions that have gone through the regulatory acceptance process.

% \Cref{fig:swiss_cheese} illustrates how the problem of aircraft collision avoidance fits into the broader context of airspace conflict management. 
Airspace conflict management relies on a layered approach to safety.
The first layer of conflict protection is strategic separation, which includes strategic airspace design and flight planning. For example, eastbound aircraft are assigned to fly at odd altitudes, while westbound aircraft are assigned to fly at even altitudes. The second layer of protection involves tactical separation techniques such as air traffic control (ATC) instructions. In less structured airspace such as the traffic pattern near an uncontrolled airport, pilots are responsible for remaining well clear of other aircraft using visual ``see and avoid'' procedures. The final layer of protection is collision avoidance, which is the focus of this article. This layer comprises systems that automatically detect impending collisions and provide advisories to resolve them. 

Collision avoidance systems are most active during the final minute before an impending collision. We call the aircraft that we can control the \textit{ownship} and the aircraft that we are trying to avoid the \textit{intruder}. The intruder may or may not have its own collision avoidance system on board. A collision avoidance system must be able to detect the intruder aircraft and determine whether a collision is imminent based on the relative geometry of the current scenario and predicted future behavior of the aircraft. If a collision is predicted, the system must then select an appropriate avoidance maneuver to resolve the conflict.

The collision avoidance problem poses a variety of technological challenges. First, the system must process data from sensors or other sources to detect and track nearby aircraft. Given these tracks, the system must understand the potential range of future behaviors of these aircraft and determine whether a collision is imminent. If a collision is predicted, the system must then select an appropriate avoidance maneuver to resolve the conflict. Importantly, the system also needs to be designed to minimize false or nuisance alerts which may distract pilots and lead to lack of trust in its advisories. These systems must be suitable for real-time flight operations and must be robust to a variety of environmental conditions. Furthermore, the system must be able to handle uncertainties in the sensor measurements, the response of the pilot, and the behavior of other aircraft.

Because aircraft collision avoidance systems are safety-critical, their designs must be subjected to a rigorous validation process. This validation process presents additional technological challenges, as it requires the system to be tested over a wide range of possible scenarios and edge cases. The results of this validation process are crucial in gaining regulatory acceptance for the system.

The technological challenges of aircraft collision avoidance systems have stimulated a multitude of productive research directions and solutions. This article dives deeper into these challenges and outlines a variety of solutions that have been studied. We place particular emphasis on solutions that have been validated and accepted by regulatory bodies. These solutions include the Airborne Collision Avoidance System X (ACAS X) \cite{Kochenderfer2012lljournal} and its predecessor, the Traffic Alert and Collision Avoidance System (TCAS) \cite{kuchar2007traffic}. These systems have been accepted as international standards.

\begin{marginnote}
\entry{ACAS X}{Airborne Collision Avoidance System X}
\entry{TCAS}{Traffic Alert and Collision Avoidance System}
\end{marginnote}

The remainder of this article is organized as follows. \Cref{sec:history} provides a brief history of aircraft collision avoidance systems. \Cref{sec:surveillance} discusses the surveillance problem, which involves detecting and tracking nearby aircraft. \Cref{sec:avoidance_maneuvers} describes the considerations involved in selecting the type of avoidance maneuvers available to an aircraft collision avoidance system. \Cref{sec:avoidance_logic} discusses ways to select collision avoidance maneuvers given tracks of nearby aircraft. Finally, \cref{sec:validation} discusses validation efforts for aircraft collision avoidance systems, and \cref{sec:regulatory_acceptance} discusses the regulatory acceptance process.

\section{HISTORY}\label{sec:history}
Today's aviation system has benefited from a remarkable transformation in safety, enabled by continual advances in air traffic control systems, training and procedures, and collision avoidance technology. As shown in \cref{fig:traffic_and_collisions}, the volume of worldwide jet traffic increased by a factor of eight between 1970 and 2025 to more than 75 million flight hours per year today. Despite this increase in traffic density, the risk of mid-air collisions for jet transport aircraft over the same period dropped by a factor of nearly one hundred to an approximate rate of 1 collision every 650 million flight hours today.

\begin{figure}
    \centering
    \begin{tikzpicture}
\begin{axis}[
  height = {7cm},
  ylabel = {\textcolor{pastelGreen}{Annual Jet Transport Flight Hours (millions)}},
  xmin = {1970},
  xmax = {2025},
  ymax = {80},
  axis y line*=right,
  axis x line=none,
  yticklabel style={pastelGreen, /pgf/number format/fixed},
  enlargelimits=false,
%   every outer y axis line/.append style={xshift=1cm},
  ymin = {0},
  width = {12cm},
  ybar
]

\addplot+[
  mark = {none},
  fill=pastelGreen!30,
  bar width=5.5pt,
  draw=none
] coordinates {
  (1970.0, 9.0)
  (1971.0, 9.5)
  (1972.0, 10.0)
  (1973.0, 10.5)
  (1974.0, 10.7)
  (1975.0, 11.0)
  (1976.0, 12.0)
  (1977.0, 12.5)
  (1978.0, 13.0)
  (1979.0, 13.7)
  (1980.0, 13.9)
  (1981.0, 13.9)
  (1982.0, 14.0)
  (1983.0, 14.5)
  (1984.0, 15.5)
  (1985.0, 16.0)
  (1986.0, 17.5)
  (1987.0, 18.5)
  (1988.0, 20.0)
  (1989.0, 21.0)
  (1990.0, 22.5)
  (1991.0, 22.8)
  (1992.0, 24.0)
  (1993.0, 25.5)
  (1994.0, 27.0)
  (1995.0, 28.5)
  (1996.0, 29.8)
  (1997.0, 31.5)
  (1998.0, 32.5)
  (1999.0, 34.3)
  (2000.0, 35.0)
  (2001.0, 34.1)
  (2002.0, 32.1)
  (2003.0, 33.9)
  (2004.0, 37.1)
  (2005.0, 40.0)
  (2006.0, 40.3)
  (2007.0, 43.0)
  (2008.0, 46.3)
  (2009.0, 45.6)
  (2010.0, 47.8)
  (2011.0, 50.9)
  (2012.0, 52.8)
  (2013.0, 54.9)
  (2014.0, 56.5)
  (2015.0, 60.0)
  (2016.0, 64.4)
  (2017.0, 67.4)
  (2018.0, 71.4)
  (2019.0, 71.9)
  (2020.0, 42.2)
  (2021.0, 46.9)
  (2022.0, 58.7)
  (2023.0, 71.5)
  (2024.0, 75.3)
  (2025.0, 75.3)
};

\node at (axis cs:1985,68) [anchor=south] {\textcolor{fignumcolor}{Mid-air Collision Rate}};
\node at (axis cs:1985,62) [anchor=south] {\textcolor{fignumcolor}{(10 year moving average)}};

\node at (axis cs:2012,36) {\textcolor{pastelGreen}{Worldwide Annual}};
\node at (axis cs:2012,30) {\textcolor{pastelGreen}{Jet Transport Flight Hours}};

\end{axis}

\begin{axis}[
  height = {7cm},
  xmin = {1970},
  xmax = {2025},
  ymax = {0.1},
  xlabel = {Year},
  yticklabel style={fignumcolor,/pgf/number format/fixed}, 
  xticklabel style={/pgf/number format/1000 sep=}, 
  enlargelimits=false,
  ymin = {0},
  width = {12cm},
  ytick pos=left,
  ylabel = {\textcolor{pastelBlue}{Midair Collision Rate (per million flight hours)}},
  xlabel = {Year},
  xticklabel shift={3pt}
]

\addplot+[
  mark = {none},
  fignumcolor,
  line width=2pt
] coordinates {
  (1970.0, 0.09709)
  (1971.0, 0.12376)
  (1972.0, 0.09921)
  (1973.0, 0.09852)
  (1974.0, 0.08380)
  (1975.0, 0.07264)
  (1976.0, 0.07709)
  (1977.0, 0.05066)
  (1978.0, 0.06623)
  (1979.0, 0.06256)
  (1980.0, 0.05993)
  (1981.0, 0.04125)
  (1982.0, 0.03994)
  (1983.0, 0.03096)
  (1984.0, 0.02985)
  (1985.0, 0.03597)
  (1986.0, 0.03460)
  (1987.0, 0.03322)
  (1988.0, 0.01905)
  (1989.0, 0.01214)
  (1990.0, 0.01153)
  (1991.0, 0.01097)
  (1992.0, 0.01560)
  (1993.0, 0.01968)
  (1994.0, 0.01862)
  (1995.0, 0.01320)
  (1996.0, 0.01252)
  (1997.0, 0.01188)
  (1998.0, 0.01132)
  (1999.0, 0.01078)
  (2000.0, 0.01031)
  (2001.0, 0.00993)
  (2002.0, 0.01289)
  (2003.0, 0.00941)
  (2004.0, 0.00912)
  (2005.0, 0.00882)
  (2006.0, 0.00855)
  (2007.0, 0.00828)
  (2008.0, 0.00798)
  (2009.0, 0.00774)
  (2010.0, 0.00750)
  (2011.0, 0.00719)
  (2012.0, 0.00457)
  (2013.0, 0.00436)
  (2014.0, 0.00418)
  (2015.0, 0.00602)
  (2016.0, 0.00383)
  (2017.0, 0.00366)
  (2018.0, 0.00350)
  (2019.0, 0.00334)
  (2020.0, 0.00338)
  (2021.0, 0.00340)
  (2022.0, 0.00168)
  (2023.0, 0.00164)
  (2024.0, 0.00159)
  (2025.0, 0.00155)
};

\addplot+[
    only marks,
    mark=*,
    mark options={fill=pastelRed, draw=none},
    mark size=2.5pt,
]
coordinates {
  (1971.0, 0.002)
  (1973.0, 0.002)
  (1976.0, 0.002)
  (1978.0, 0.002)
  (1979.0, 0.002)
  (1985.0, 0.002)
  (1986.0, 0.002)
  (1992.0, 0.002)
  (1993.0, 0.002)
  (1996.0, 0.002)
  (2002.0, 0.002)
  (2006.0, 0.002)
  (2012.0, 0.002)
  (2015.0, 0.002)
  (2025.0, 0.002)
};

\end{axis}
\end{tikzpicture}
    \caption{\label{fig:traffic_and_collisions} Worldwide jet transport flight hours and mid-air collision rates from 1970--2025. Mid-air collision events are indicated with red markers derived from data provided by \citeauthor{BoeingAccidentSummary2025} \cite{BoeingAccidentSummary2025}.}
\end{figure}

Modern aviation collision avoidance systems are built upon a foundation of technologies that have evolved over the last century. The earliest years of flight depended entirely on pilot ``see and avoid'' at close range. This approach quickly proved to be insufficient with the first recorded mid-air collision in 1910 \cite{CookEuroATM2007}. The development of airborne radio in the 1930s enabled pilots to report their position and provided a rudimentary method to coordinate and avoid traffic at distances beyond visual range. Following World War II, primary radar allowed air traffic controllers to track aircraft without relying on verbal position reports, albeit only by referencing luminescent blips on a screen. 

The subsequent development of secondary radar systems that use transponders on board provided aircraft identity information in the air traffic controller's data block and, starting in the 1970s, altitude. This additional information in turn enabled ground-based alerting capabilities to warn air traffic controllers of projected traffic conflicts based on their current trajectories. Enhancement continued with the development of the Mode Select (Mode S) transponder in the 1980s. Mode S enabled ground radars and, importantly, aircraft to interrogate and track individual aircraft in a way that accommodated the rapid growth in aircraft density in busy terminal areas.

Two major mid-air collisions in the United States, above San Diego, CA in 1978 and Cerritos, CA in 1986, led to mandates requiring large aircraft to equip with an independent collision avoidance system that would operate in parallel to air traffic control and provide guidance directly to flight crews. This functionality required the ability for aircraft to detect and track one another, a function provided by the concurrently-developed Mode S transponder. Also enabled by Mode S was a datalink in which additional information, such as the directionality of a collision avoidance maneuver, could be communicated between aircraft. This same datalink is the core behind today's Automatic Dependent Surveillance-Broadcast (ADS-B) system in which aircraft periodically report their position and additional state information to receivers on the ground or in the air and thus do not rely on ground radar.

\begin{marginnote}
\entry{ADS-B}{Automatic Dependent Surveillance-Broadcast}
\end{marginnote}

Today's currently-mandated airborne collision avoidance system, TCAS, is required in the United States on all turbine-powered aircraft with more than 30 seats or with a maximum takeoff weight above \SI{15000}{\kilogram} and by the International Civil Aviation Organization on aircraft with more than 19 seats or more than \SI{5700}{\kilogram} \cite{FAATCASII2011}. Furthermore, many other aircraft, such as business jets and rotorcraft, have voluntarily equipped with TCAS to benefit from the system.

% TCAS uses the Mode S datalink and a directional antenna to track nearby aircraft using a linear trajectory projection, display their relative position on a planform flight deck display, and generate Traffic Advisory caution alerts to cue pilots to visually acquire nearby traffic. Should a conflict worsen, TCAS subsequently generates a Resolution Advisory (RA) which directs the flight crew to climb, descend, or maintain or modify their vertical rate. TCAS RAs are coordinated between TCAS-equipped aircraft to ensure they maneuver in complementary directions.

Since its broad deployment starting in the mid-1990s, TCAS has been cited with preventing a number of potential mid-air collisions and is widely considered to be an important contributor to the safety improvements over the last three decades shown in \cref{fig:traffic_and_collisions}. A continuing challenge, however, is that the design of TCAS is tightly coupled to Mode S surveillance and assumptions about aircraft performance and airspace densities that were appropriate in the 1990s but which have changed in the intervening decades. One key impact of these constraints is that TCAS has been prone to issuing a high rate of nuisance alerts as traffic densities have increased, especially in dense terminal areas where there may be a mix of traffic using visual separation procedures \cite{folmar1994extension,chamlou2009future}.

Modifying the TCAS algorithms to reduce nuisance alerts while maintaining safety has proven to be a significant challenge \cite{RTCADO3372012}. TCAS was designed using a classical rule-based, expert-system approach, with numerous if-then logical statements operating on simple linear trajectory projections of each aircraft in a potential conflict. Any change to the logic or its parameters can have far-reaching impacts on the resulting collision risk and alerting rate that are difficult to predict. The resulting incremental, iterative design challenge means that even small changes to TCAS require extensive analysis before they can be approved for deployment.

With the advent of modern machine learning techniques and growth in computational power, alternate approaches to the design of TCAS were proposed in the 2000s in which alerting decisions would be derived from probabilistic encounter characteristics rather than through traditional expert-system approaches \cite{kuchar2002review,yang2002performance,winder2004hazard,wolf2011aircraft,Temizer2010}.
This earlier work ultimately led to the development of the next generation airborne collision alerting system, ACAS X \cite{Kochenderfer2012lljournal}. The intent of the ACAS X design approach was to derive an optimal alerting policy that balances safety, nuisance alerts, and other operational acceptability considerations. It also allows for the flexible integration of a variety of surveillance technologies (e.g., Mode S interrogation, ADS-B, airborne radar, electro-optical sensors) and can be adapted to a range of aircraft types and performance characteristics including small and large UAS and rotorcraft. As our airspace continues to evolve, aircraft collision avoidance systems will need to continue to adapt to new challenges.

\section{SURVEILLANCE}\label{sec:surveillance}
The first step in avoiding mid-air collisions is to detect nearby aircraft. This step requires a robust surveillance system that can accurately track the state of all aircraft in close proximity. 
%The primary surveillance sources used in aviation today include transponders, Automatic Dependent Surveillance-Broadcast (ADS-B), and other various onboard sensors.
As noted in \cref{sec:history}, the first surveillance sources used for collision avoidance were transponders, which automatically send and reply to interrogation signals by transmitting coded information such as identity. The reply message also includes altitude information, which is critical for determining potential conflicts. However, this altitude information is subject to quantization errors, which must be accounted for in the surveillance system \cite{Asmar2013}. Range information can be derived from the time it takes to receive a reply to an interrogation. Bearing information can be estimated from signal amplitude and phase differences, but this estimation is difficult and results in time-correlated errors \cite{Panken2014}.

When TCAS was originally developed, transponder-based surveillance was the primary means of detecting nearby aircraft. TCAS interrogators send out interrogation messages at a \SI{0.2}{\hertz} or \SI{1}{\hertz} rate, depending on the current threat level. TCAS uses the 1030/1090 MHz frequency bands, which are also used by ATC ground sensors. Therefore, interrogation rates must be selected to ensure minimal impact on the 1030 MHz channel \cite{kuchar2007traffic}. 

The introduction of ADS-B in the late 1990s and early 2000s provided an additional surveillance source based on data from the Global Navigation Satellite System (GNSS) that could improve performance \cite{romli2008impact}. ADS-B must still be validated by transponder interrogation before it can be used for collision avoidance, but it can significantly decrease the required number of interrogations \cite{RTCADO3002006}. The use of advanced decision-making techniques to decide when to validate ADS-B information reduces this frequency even further \cite{Kochenderfer2011dasc}.

The information provided by transponders, ADS-B, and other onboard sensors is sufficient to support the collision avoidance of transponder-equipped aircraft. However, small aircraft such as gliders and ultralights may not be equipped with transponders and require other surveillance sources to detect. For example, air-to-air radar and electro-optical/infrared (EO/IR) sensors can be used \cite{Griffith2008,Smyers2023}. While these sensors can be difficult to model, it is important to assess their detection capabilities and derive requirements for them \cite{Griffith2008}. Surrogate modeling and adaptive sampling techniques can increase the efficiency of this process \cite{jones2018surrogate,Katz2023dasc}. ACAS X is designed to accommodate a variety of surveillance sources, including those used for non-transponder aircraft \cite{Kochenderfer2012lljournal}.

The surveillance and tracking module of a collision avoidance system must integrate surveillance information both across time and across different surveillance sources. First, state estimation techniques such as Kalman filtering are applied over time to create tracks for each surveillance source \cite{bar2001estimation}. Next, the system must aggregate information across each track at the current time step. To determine which tracks correspond to the same target, ACAS X uses aircraft identification or a distance metric when no information is present. The information from the best surveillance source for each target is selected for use in the collision avoidance algorithms.

The surveillance system passes information to the threat resolution portion of the aircraft collision avoidance system in the form of either a point estimate or a distribution. The TCAS decision-making logic operates on point estimates. In contrast, ACAS X achieves improved performance by incorporating uncertainty information into its decision-making process \cite{Chryssanthacopoulos2011jgcd}. In particular, it operates on a belief state representation involving a finite set of weighted samples \cite{julier1997new}.

\section{AVOIDANCE MANEUVERS}\label{sec:avoidance_maneuvers}
Once an aircraft collision avoidance system has detected an aircraft, it must warn pilots of the traffic and select an appropriate avoidance maneuver. Both TCAS and ACAS X issue two types of warnings. The first type is referred to as a traffic alert (TA), which warns pilots of nearby traffic and aids with visual acquisition \cite{Puntin2013}. If the threat of nearby traffic becomes high enough to suggest a potential impending collision, the system issues a second type of advisory known as a resolution advisory (RA), which recommends a specific avoidance maneuver to the pilot. 
\begin{marginnote}
\entry{TA}{Traffic Alert}
\entry{RA}{Resolution Advisory}
\end{marginnote}

TCAS and ACAS X select these RAs from a set of available maneuvers. The types of avoidance maneuvers available are dependent on a variety of factors including aircraft type, airspace structure, and operational suitability. For example, a small unmanned aircraft may not be able to perform the same avoidance maneuvers as a large commercial airliner. Furthermore, maneuvers should be kept simple such that pilots can easily understand and execute them in high-stress situations \cite{kuchar2007traffic}.

Because commercial aircraft typically have better vertical maneuverability than horizontal maneuverability, TCAS was designed to issue vertical advisories. Vertical maneuvers also tend to be less disruptive since they allow the aircraft to remain on its horizontal flight path. TCAS RAs represent target vertical rate ranges. For example, an RA of \textsc{DES1500} indicates that the aircraft should descend at a rate of at least \SI{1500}{\foot\per\minute}. Advisories may also require the aircraft to strengthen its current climb or descent, maintain its current vertical rate, or level off. ACAS Xa, the version of ACAS X designed to replace TCAS, selects from the same set of vertical advisories \cite{Kochenderfer2012lljournal}.

Collision avoidance systems designed for smaller or less maneuverable aircraft require a different set of avoidance maneuvers. For example, the versions of ACAS X designed for unmanned aircraft (ACAS Xu, ACAS sXu, and ACAS Xr) select from a restricted set of vertical advisories with lower vertical rate targets. To make up for this decreased maneuverability, they also issue horizontal advisories \cite{owen2019acas,alvarez2019acas}. Horizontal advisories represent target turn rate ranges, which are typically expressed in degrees per second. Horizontal and vertical advisories may be issued simultaneously or selected dynamically based on the current situation \cite{Owen2016}.

New aircraft types such as quadrotors and electric vertical takeoff and landing (eVTOL) vehicles have different maneuverability characteristics than traditional fixed-wing aircraft. These vehicles have greater ability to change their speed or hover in place. As a result, recent studies have explored the efficacy of speed change advisories in collision avoidance systems designed for these aircraft types \cite{Mueller2016pomdp,Katz2022aviation}. However, these studies found limited benefit to speed change advisories in the collision avoidance time horizon but that they may have more utility in providing separation further out in time.

\section{THREAT RESOLUTION LOGIC}\label{sec:avoidance_logic}
The \textit{threat resolution logic} of an aircraft collision avoidance system is responsible for selecting an appropriate collision avoidance maneuver from the set of available maneuvers given the current state estimate. Several categories of solutions have been proposed based on ideas from geometric reasoning \cite{ikeda2002automatic,kuchar2007traffic,kim2007uav,luongo2009optimal,luongo2011automatic,albaker2011autonomous,lin2020fast}, control theory~\cite{squires2018constructive,liu2019aircraft,molnar2025collision}, artificial potential fields \cite{du2019real}, path planning \cite{lin2015collision,lin2016sampling,lin2017sampling}, and dynamic programming~\cite{Kochenderfer2010itsc,Temizer2010,Kochenderfer2011atc371,Kochenderfer2012lljournal,Kochenderfer2012atcq,Kochenderfer2013,Sunberg2016,Mueller2016pomdp,owen2019acas,alvarez2019acas,bertram2022fast,Katz2022aviation}. These solutions present a variety of trade-offs in terms of performance, computational complexity, operational suitability, and robustness to uncertainties \cite{Mueller2016comp}. For example, while geometric approaches often provide fast, analytical solutions, they may not be as robust to uncertainties as dynamic programming approaches. Furthermore, artificial potential field and path planning approaches typically produce collision avoidance maneuvers that require precise control over the aircraft trajectory, which may be difficult for pilots to execute. Finally, open-loop solutions, such as those based on geometric reasoning and traditional path planning methods, have been shown to perform poorly in highly stochastic environments when compared to closed-loop dynamic programming solutions \cite{Chryssanthacopoulos2011itsc}. The remainder of the section discusses the solutions used in TCAS and ACAS X, which fall into the categories of geometric reasoning and dynamic programming, respectively.

\subsection{TCAS Logic}
The TCAS threat resolution logic is based on a collection of heuristic rules. First, the logic extrapolates the ownship and intruder trajectories assuming constant velocity. If the time to closest point of approach and projected miss distance are below certain thresholds, the system decides to issue a TA or RA. If this check indicates that an RA is required, the system selects a maneuver based on the relative state of the ownship and intruder.

Selecting an RA involves two main steps. \Cref{fig:tcas_logic} demonstrates these steps in an example scenario. First, TCAS projects the result of applying a standard climb maneuver and a standard descend maneuver. It assumes a five second pilot response delay followed by a constant vertical acceleration. The system then selects the \textit{sense} for the RA by choosing the vertical direction of the projected maneuver that results in the largest vertical separation at the time of closest approach. 

\begin{marginnote}
\entry{Sense}{direction an advisory instructs a pilot to maneuver (either \textit{up} or \textit{down}).}
\end{marginnote}

The sense determines the set of advisories to consider in the next step. For example, if the selected sense is \textit{down}, the system considers advisories such as limiting a current climb, leveling off from a climb, or decending at a particular vertical rate. In the second step, TCAS projects the result of applying each of the advisories consistent with the selected sense. Each projection again assumes a five second pilot response delay followed by a constant vertical acceleration. Based on the results, TCAS selects the minimum-strength RA that results in a minimum required vertical separation at the time of closest approach.

\begin{figure} % tcas logic
\begin{center}
\begin{tikzpicture}
    % Step 1 box in upper left
    \node [draw=fignumcolor, fill=white, rounded corners=2pt, thick] at (-1.5, 1.5) {\small \textcolor{fignumcolor}{\textbf{step 1}}};
    \draw[fignumcolor, line width=1pt] (-0.9, 1.5) -- (10, 1.5);
    
    \draw[fignumcolor] (-1.969615506,-1.3472963553) -- (0,-1);
    \draw[rounded corners, densely dashed, black!60, -{Triangle[fill=black!60,length=2.5mm]}] (0,-1) -- (1,-0.8236730193) -- (3.5, 1.1970529229);
    \draw[rounded corners, densely dashed, pastelGreen, -{Triangle[fill=pastelGreen,length=2.5mm]}] (0,-1) -- (1,-0.8236730193) -- (3.5, -1.4408174518);

    \draw[fignumcolor] (9, 0) -- (7, 0);
    \draw[densely dashed, fignumcolor, -{Triangle[fill=fignumcolor,length=2.5mm]}] (7,0) -- (3.5, 0);

    \draw[<->, black!60, line width=0.75pt] (3.5, 0.05) -- (3.5, 1.1470529229);
    \draw[<->, pastelGreen, line width=0.75pt] (3.5, -0.05) -- (3.5, -1.3908174518);

    \node [aircraft side,fill=black,draw=black,minimum width=1cm,rotate=10,scale=1.2] at (0,-1) {};
    \node [aircraft side,fill=black,draw=black,minimum width=1cm,rotate=0,scale=1.2,xscale=-1] at (7,0) {}; 
    \node [below] at (0,-1.3) {\footnotesize ownship};
    \node [below] at (7,-0.3) {\footnotesize intruder};

    \node [black!60, anchor=west] at (3.6, 0.65) {\small result of climb};
    \node [pastelGreen, anchor=west] at (3.6, -0.75) {\small result of descent};

    % % Wide outlined arrow with transparent fill pointing down
    % \draw[thick, fill=white, fill opacity=0.3, fill=black!40] (3.35, -2.0) -- (3.35, -2.9) -- (3.2, -2.9) -- (3.5, -3.2) -- (3.8, -2.9) -- (3.65, -2.9) -- (3.65, -2.0) -- cycle;
    % \node [anchor=west] at (3.8, -2.45) {\small descend sense};
\end{tikzpicture}

\vspace{5mm}

\begin{tikzpicture}
    % Step 1 box in upper left
    \node [draw=fignumcolor, fill=white, rounded corners=2pt, thick] at (-1.5, 1.5) {\small \textcolor{fignumcolor}{\textbf{step 2}}};
    \draw[fignumcolor, line width=1pt] (-0.9, 1.5) -- (10, 1.5);

    % Wide outlined arrow with transparent fill pointing down
    \draw[thick, fill=white, fill opacity=1.0, fill=black!40] (3.35, 2.3) -- (3.35, 1.4) -- (3.2, 1.4) -- (3.5, 1.1) -- (3.8, 1.4) -- (3.65, 1.4) -- (3.65, 2.3) -- cycle;
    \node [anchor=west] at (3.7, 1.85) {\small down sense};

    \fill[pastelRed!30] (3.5, -1.25) rectangle (3.75, 0.0);
    \fill[pastelGreen!30] (3.5, -3.0) rectangle (3.75, -1.25);
    \draw[black!60] (3.5, -3.0) -- (3.5, 0.0);
    
    % Curly brace for red rectangle
    \draw [decorate, decoration={brace, amplitude=5pt}] (3.75, 0.0) -- (3.75, -1.25);
    \node [anchor=west] at (3.9, -0.625) {\small 400 ft};
    
    \draw[fignumcolor] (-1.969615506,-1.3472963553) -- (0,-1);
    \draw[rounded corners, densely dashed, pastelRed, -{Triangle[fill=pastelRed,length=2.5mm]}] (0,-1) -- (1,-0.8236730193) -- (3.5, -0.2065285868);
    \draw[rounded corners, densely dashed, pastelRed, -{Triangle[fill=pastelRed,length=2.5mm]}] (0,-1) -- (1,-0.8236730193) -- (3.5, -0.8236730193);
    \draw[rounded corners, pastelGreen, -{Triangle[fill=pastelGreen,length=2.5mm]}, line width=1pt] (0,-1) -- (1,-0.8236730193) -- (3.5, -2.0975688392);
    \draw[rounded corners, densely dashed, pastelGreen, -{Triangle[fill=pastelGreen,length=2.5mm]}] (0,-1) -- (1,-0.8236730193) -- (3.5, -2.8443989615);

    % Labels for trajectory lines
    \node [pastelRed] at (2.5, -0.15) {\scriptsize \textsc{DNC500}};
    \node [pastelRed] at (2.5, -0.65) {\scriptsize \textsc{DNC}};
    \node [pastelGreen] at (2.5, -2.0) {\scriptsize \textsc{DES1500}};
    \node [pastelGreen] at (2.5, -2.5) {\scriptsize \textsc{DES2500}};

    \draw[fignumcolor] (9, 0) -- (7, 0);
    \draw[densely dashed, fignumcolor, -{Triangle[fill=fignumcolor,length=2.5mm]}] (7,0) -- (3.5, 0);

    \node [aircraft side,fill=black,draw=black,minimum width=1cm,rotate=10,scale=1.2] at (0,-1) {};
    \node [aircraft side,fill=black,draw=black,minimum width=1cm,rotate=0,scale=1.2,xscale=-1] at (7,0) {}; 
    \node [below] at (0,-1.3) {\footnotesize ownship};
    \node [below] at (7,-0.3) {\footnotesize intruder};
\end{tikzpicture}
\end{center}
\caption{\label{fig:tcas_logic} Simplified TCAS logic for an example scenario with a minimum required separation of \SI{400}{\foot}. In the first step, the logic selects between a climb or descend sense by projecting the results of two maneuver templates assuming a five second pilot response delay and selecting the sense of the maneuver (in this case, down) that results in the greatest vertical separation from the intruder. In the second step, the logic projects the results of four different vertical RAs (limit climb to \SI{500}{\foot\per\minute}, do not climb, descend at \SI{1500}{\foot\per\minute}, and descend at \SI{2500}{\foot\per\minute}). It then selects the minimum-strength RA that results in at least \SI{400}{\foot} of vertical separation from the intruder. In this case, the logic will advise the pilot to descend at \SI{1500}{\foot\per\minute} (shown in solid green).}
\end{figure}

These two steps provide a simplified explanation of the TCAS logic. In practice, many heuristic rules and parameters are required to ensure the system is robust to sensor uncertainty, operationally suitable, and safe in the wide range of scenarios it may encounter in the airspace. For example, if TCAS is highly uncertain about the track of an intruder aircraft, it will refrain from issuing an RA. Other rules specify when TCAS should strengthen, weaken, or reverse an RA.

Prior modeling and simulation analysis estimates that TCAS reduces the risk of mid-air collision by a factor of approximately \num{8} if one aircraft is equipped with the system and by a factor of approximately \num{62} if both aircraft are equipped with the system and respond to their RAs \cite{EspindleGriffithKucharATC3492009}. However, the open-loop, deterministic nature of the TCAS logic can result in undesirable outcomes in scenarios where behavior differs significantly from its internal assumptions. In fact, this drawback was a factor in a mid-air collision over \"{U}berlingen, Germany in 2002 where the pilots of one aircraft followed the instructions from ATC, which conflicted with their TCAS advisory. Furthermore, the design of TCAS also began to result in excessive alerting as airspace density increased, but due to the complex rules embedded in the logic, TCAS is difficult to adapt to new aircraft types and airspace structures. For this reason, the FAA began supporting the development of a next generation aircraft collision avoidance system at the beginning of the 2010s, which resulted in the development of ACAS~X.

\subsection{ACAS X Logic}
Instead of representing the threat resolution logic as a set of heuristic rules, ACAS X represents the logic as an optimized numeric table. The table maps the current state of the ownship and intruder aircraft to a cost for each available avoidance maneuver. The logic selects the maneuver with the lowest cost. This representation has multiple advantages over the TCAS representation. For instance, implementing and updating the ACAS X logic is more straightforward for manufacturers. Rather than requiring manufacturers to translate a complex set of rules into code, they can instead perform lookups in the provided logic table. Moreover, updates to the logic simply require uploading a new version of the table. 

The ACAS X logic also better accounts for different types of uncertainty than the TCAS logic. Specifically, the optimization and implementation of the ACAS X logic table takes into account the two main types of uncertainty present in the collision avoidance problem (\cref{fig:state_vs_outcome}). The first type of uncertainty is outcome uncertainty, which refers to the uncertainty in the future behavior of the ownship and intruder aircraft. The second type of uncertainty is state uncertainty, which refers to the uncertainty in the current state estimate of the ownship and intruder aircraft caused by sensor noise and other imperfect information. ACAS X accounts for outcome uncertainty during the optimization of the logic table and accounts for state uncertainty during the table lookup process.

\begin{figure}[t] % state vs outcome
\begin{center}
\begin{tikzpicture}
    \draw[rounded corners] (-1, -2) rectangle (4.5, 1.5);
    \draw[rounded corners] (5, -2) rectangle (11.5, 1.5);

    \draw [fignumcolor, rounded corners, densely dashed] (0,-1) -- (0.6, -0.8942038116) -- (1.7, 0.0707607643);
    \draw [fignumcolor, rounded corners, densely dashed] (0,-1) -- (0.6, -0.8942038116) -- (1.7, -0.230251002);
    \draw [fignumcolor, rounded corners, line width=1pt] (0,-1) -- (0.6, -0.8942038116) -- (1.7, -0.4756380422);
    \draw [fignumcolor, rounded corners, densely dashed] (0,-1) -- (0.6, -0.8942038116) -- (1.7, -0.6914277838);
    \draw [fignumcolor, rounded corners, densely dashed] (0,-1) -- (0.6, -0.8942038116) -- (1.7, -0.8942038116);

    \draw [fignumcolor, line width=1pt] (3.5, 0) -- (1.8, 0);
    \draw [fignumcolor, rounded corners=10pt, densely dashed] (3.5, 0) -- (2.9, 0) -- (1.8, 0.2027760278);
    \draw [fignumcolor, rounded corners=10pt, densely dashed] (3.5, 0) -- (2.9, 0) -- (1.8, -0.2027760278);
    \draw [fignumcolor, rounded corners=10pt, densely dashed] (3.5, 0) -- (2.9, 0) -- (1.8, 0.4185657694);
    \draw [fignumcolor, rounded corners=10pt, densely dashed] (3.5, 0) -- (2.9, 0) -- (1.8, -0.4185657694);

    \node [aircraft side,fill=black,draw=black,minimum width=1cm,rotate=10,scale=1] at (0,-1) {};
    \node [aircraft side,fill=black,draw=black,minimum width=1cm,rotate=0,scale=1,xscale=-1] at (3.5,0) {}; 
    \node [below] at (0,-1.3) {\footnotesize ownship};
    \node [below] at (3.5,-0.3) {\footnotesize intruder};
    
    \node [aircraft side,fill=black,draw=black,minimum width=1cm,rotate=10,scale=1] at (6.5,-1) {};
    \node [aircraft side,fill=black,draw=black,minimum width=1cm,rotate=0,scale=1,xscale=-1] at (10,0) {}; 
    \node [below] at (6.5,-1.3) {\footnotesize ownship};
    \node [below] at (10,-0.3) {\footnotesize intruder};

    \draw [fignumcolor!20, line width=1pt] (6.35,-1) ellipse (1.0 and 0.5);
    \draw [fignumcolor!40, line width=1pt] (6.35,-1) ellipse (0.8 and 0.4);
    \draw [fignumcolor!60, line width=1pt] (6.35,-1) ellipse (0.6 and 0.3);
    \draw [fignumcolor!80, line width=1pt] (6.35,-1) ellipse (0.4 and 0.2);
    \draw [fignumcolor!20, line width=1pt] (10.1,0) ellipse (1.0 and 0.5);
    \draw [fignumcolor!40, line width=1pt] (10.1,0) ellipse (0.8 and 0.4);
    \draw [fignumcolor!60, line width=1pt] (10.1,0) ellipse (0.6 and 0.3);
    \draw [fignumcolor!80, line width=1pt] (10.1,0) ellipse (0.4 and 0.2);
    \fill[fignumcolor] (6.35, -1) circle (0.08);
    \fill[fignumcolor] (10.1, 0) circle (0.08);

    \node at (1.75, 1.0) {Outcome Uncertainty};
    \node at (8.25, 1.0) {State Uncertainty};
\end{tikzpicture}
\end{center}
\caption{\label{fig:state_vs_outcome} Two types of uncertainty address by ACAS X. The left side shows the outcome uncertainty, which is addressed during the optimization of the logic table. The right side shows the state uncertainty, which is addressed during the lookup process.}
\end{figure}

\subsubsection{Table Optimization}
The costs in the ACAS X logic table are determined using an optimization technique called \textit{dynamic programming}. The collision avoidance problem is first formulated as a Markov decision process (MDP), which accounts for outcome uncertainty \cite{Kochenderfer2022}. An MDP is a mathematical framework for modeling sequential decision-making problems that consists of a state space, an action space, a transition model, and a reward function. To model the collision avoidance problem as an MDP, we must define each of these elements.
\begin{marginnote}
\entry{MDP}{Markov decision process}
\end{marginnote}

The state space for the collision avoidance problem is the set of all possible states that the system may encounter, and the action space is the discrete set of available RAs. ACAS X uses several variables to describe the current state of the ownship and intruder aircraft (\cref{fig:mdp_states}). For the variants of ACAS X that issue vertical RAs, the state consists of the relative altitude between the ownship and intruder aircraft $h$, the ownship vertical rate $\dot h_0$, the intruder vertical rate $\dot h_1$, the previous advisory $a_\text{prev}$, and the time to loss of horizontal separation $\tau$ \cite{Kochenderfer2010itsc}. For the variants of ACAS X that issue horizontal RAs, the state consists of the range $r$, the bearing $\theta$, the relative heading $\psi$, the ownship speed $v_0$, the intruder speed $v_1$, the previous advisory $a_\text{prev}$, and the time to loss of vertical separation $\tau$ \cite{owen2019acas}.

\begin{figure}[t]
\begin{center}
\begin{tikzpicture}
    \draw[rounded corners] (-1, -2) rectangle (4.5, 2.2);
    \draw[rounded corners] (5, -2) rectangle (11.5, 2.2);

    %%%%%%%%%%%% Vertical %%%%%%%%%%%%%
    \draw[black!60, ->] (0, -1) -- (1, -0.5336923418);
    \draw[black!60, ->] (3.5, 0) -- (2.5, 0.4663076582);
    \draw[dotted] (-1, -1) -- (4.5, -1);
    \draw[dotted] (-1, 0) -- (4.5, 0);
    \draw[fignumcolor, -stealth, line width=1pt] (1, -1) -- (1, -0.5336923418);
    \node[fignumcolor, anchor=west] at (1.04, -0.75) {\footnotesize $\dot h_0$};
    \draw[fignumcolor, -stealth, line width=1pt] (2.5, 0) -- (2.5, 0.4663076582);
    \node[fignumcolor, anchor=east] at (2.48, 0.25) {\footnotesize $\dot h_1$};

    \draw[fignumcolor, stealth-stealth, line width=1pt] (1.75, -1) -- (1.75, 0);
    \node[fignumcolor, anchor=west] at (1.78, -0.5) {\footnotesize $h$};

    % \draw[fignumcolor, stealth-stealth, line width=1pt] (0, -1) -- (1.75, -1);
    % \node[fignumcolor] at (0.875, -1.2) {\footnotesize $\tau$};

    \node [aircraft side,fill=black,draw=black,minimum width=1cm,rotate=25,scale=1] at (0,-1) {};
    \node [aircraft side,fill=black,draw=black,minimum width=1cm,rotate=-25,scale=1,xscale=-1] at (3.5,0) {}; 
    \node [below] at (0,-1.3) {\footnotesize ownship};
    \node [below] at (3.5,-0.3) {\footnotesize intruder};

    %%%%%%%%%%%% Horizontal %%%%%%%%%%%%%
    \draw[fignumcolor, line width=1pt] (6, -1) -- (9.5, 0.25);
    \node[fignumcolor, below] at (7.75,-0.375) {\footnotesize $r$};

    \draw[fignumcolor, -stealth, line width=1pt] (6, -1) -- (7.25, -1);
    \node[fignumcolor, anchor=west] at (7.26, -1) {\footnotesize $v_0$};

    \draw[fignumcolor, -stealth, line width=1pt] (9.5, 0.25) -- (8.6161165235, 1.1338834765);
    \node[fignumcolor] at (8.6, 1.3) {\footnotesize $v_1$};

    \draw[fignumcolor, line width=0.75pt] (6.6,-1) arc (0:20:0.6);
    \node[fignumcolor, anchor=west] at (6.7,-0.85) {\footnotesize $\theta$};

    \draw[dotted] (9.5, 0.25) -- (10.75, 0.25);
    \draw[fignumcolor, line width=0.75pt] (10.0,0.25) arc (0:135:0.5);
    \node[fignumcolor] at (10.0, 0.9) {\footnotesize $\psi$};

    \node [aircraft top,fill=black,draw=black,minimum width=1cm,rotate=0,scale=0.8] at (6,-1) {};
    \node [aircraft top,fill=black,draw=black,minimum width=1cm,rotate=135,scale=0.8] at (9.5,0.25) {};
    \node [below] at (6,-1.3) {\footnotesize ownship};
    \node [below] at (9.5,-0.3) {\footnotesize intruder};
    
    \node at (1.75, 1.8) {Vertical Logic State};
    \node at (8.25, 1.8) {Horizontal Logic State};
\end{tikzpicture}
\end{center}
\caption{\label{fig:mdp_states} State variables used in the ACAS X MDP. The vertical logic state consists of the relative altitude $h$, ownship vertical rate $\dot h_0$, intruder vertical rate $\dot h_1$, the previous advisory (not shown), and the time to loss of horizontal separation $\tau$. The horizontal logic state consists of the range $r$, bearing $\theta$, relative heading $\psi$, ownship horizontal speed $v_0$, intruder horizontal speed $v_1$, the previous advisory (not shown), and the time to loss of vertical separation $\tau$.}
\end{figure}

To ensure that the dynamic programming solution remains computationally tractable, it is important to limit the dimensionality of the state space. Both the vertical and horizontal states maintain this property by wrapping all uncontrolled state variables into a single variable $\tau$. For example, because the vertical logic only has control over the vertical state variables, all horizontal variables are summarized by the time to loss of horizontal separation~$\tau$. This formulation is consistent with the partially controlled Markov decision process framework introduced by \citeauthor{Kochenderfer2013} \cite{Kochenderfer2013} and allows for efficient optimization.

The transition model of an MDP is a probabilistic model representing the probability of transitioning to a state $s^\prime$ given that we take action $a$ in state $s$. For the collision avoidance problem, this model takes into account the outcome uncertainty due to lack of information about the intruder aircraft trajectory, pilot response delays, and other environment factors. For example, when determining the likelihood of each possible next state, the intruder is assumed to apply random accelerations according to a zero-mean Gaussian distribution. Moreover, the transition model incorporates a probabilistic model of pilot response delay using a geometric distribution \cite{Chryssanthacopoulos2011acc}.

The reward function for ACAS X uses a variety of penalties that balance between safety and operational efficiency. For example, a large cost is incurred for reaching collision states, while a smaller cost is incurred for issuing an RA, strengthening an RA, and reversing an RA. Optimizing the collision avoidance logic to satisfy a reward function with this structure results logic that allows for safe operation without issuing unnecessary advisories. However, the parameters of the reward function must be carefully tuned to strike the right balance between safety and operational efficiency. Surrogate modeling allows for efficient tuning of the parameters \cite{Holland2013}.

\begin{textbox}[h]
\section{DYNAMIC PROGRAMMING SOLUTION}
% \vspace{3mm}
The figure below (adapted from \citeauthor{Kochenderfer2012lljournal} \cite{Kochenderfer2012lljournal}) shows a notional example of the dynamic programming solution to the collision avoidance problem. The vertical axis represents the relative altitude between the ownship and intruder aircraft, and the horizontal axis represents time. The box in the center of the rightmost column represents a collision state.
\begin{center}
\begin{tikzpicture}
    \node at (0,0) {
    \begin{tikzpicture}[x=0.9cm,y=0.9cm]
        \casgrid
        \colorit{pastelRed}{3,2}
        \colorit{pastelGreen!50}{3,1}
        \colorit{pastelGreen}{3,0}
        \colorit{pastelGreen!50}{3,3}
        \colorit{pastelGreen}{3,4}
        \threesplit{2,2};
        \draw[-stealth, line width=0.5pt] (2.35, 2.3) -- (2.65, 2.3);
        \draw[-stealth, line width=1pt] (2.35, 2.3) -- (2.65, 2.65);
        \draw[-stealth, line width=0.5pt] (2.35, 2.3) -- (2.65, 3.65);
        \fill[white, draw=black, line width=0.5pt] (-0.75, 4.25) rectangle (-0.25, 4.75);
        \node at (-0.5, 4.5) {\footnotesize $1$};
    \end{tikzpicture}
    };

    \node at (5,0) {
    \begin{tikzpicture}[x=0.9cm,y=0.9cm]
        \casgrid
        \colorit{pastelRed}{3,2}
        \colorit{pastelGreen!50}{3,1}
        \colorit{pastelGreen}{3,0}
        \colorit{pastelGreen!50}{3,3}
        \colorit{pastelGreen}{3,4}
        \threesplitcolor{2,2}{pastelGreen!30}{pastelRed!50}{pastelGreen!30}
        \fill[white, draw=black, line width=0.5pt] (-0.75, 4.25) rectangle (-0.25, 4.75);
        \node at (-0.5, 4.5) {\footnotesize $2$};
    \end{tikzpicture}
    };

    \node at (10,0) {
    \begin{tikzpicture}[x=0.9cm,y=0.9cm]
        \casgrid
        \colorit{pastelRed}{3,2}
        \colorit{pastelGreen!50}{3,1}
        \colorit{pastelGreen}{3,0}
        \colorit{pastelGreen!50}{3,3}
        \colorit{pastelGreen}{3,4}
        \labeldescend{pastelGreen!30}{2,2}
        \fill[white, draw=black, line width=0.5pt] (-0.75, 4.25) rectangle (-0.25, 4.75);
        \node at (-0.5, 4.5) {\footnotesize $3$};
    \end{tikzpicture}
    };

    \node at (0,-6) {
    \begin{tikzpicture}[x=0.9cm,y=0.9cm]
        \casgrid
        \colorit{pastelRed}{3,2}
        \colorit{pastelGreen!50}{3,1}
        \colorit{pastelGreen}{3,0}
        \colorit{pastelGreen!50}{3,3}
        \colorit{pastelGreen}{3,4}
        \labeldescend{pastelGreen!30}{2,2}
        \labeldescend{pastelGreen!50}{2,1}
        \labeldescend{pastelGreen!75}{2,0}
        \labelclimb{pastelGreen!50}{2,3}
        \labelclimb{pastelGreen!75}{2,4}
        \fill[white, draw=black, line width=0.5pt] (-0.75, 4.25) rectangle (-0.25, 4.75);
        \node at (-0.5, 4.5) {\footnotesize $4$};
    \end{tikzpicture}
    };

    \node at (5,-6) {
    \begin{tikzpicture}[x=0.9cm,y=0.9cm]
        \casgrid
        \colorit{pastelRed}{3,2}
        \colorit{pastelGreen!50}{3,1}
        \colorit{pastelGreen}{3,0}
        \colorit{pastelGreen!50}{3,3}
        \colorit{pastelGreen}{3,4}
        \labeldescend{pastelGreen!30}{2,2}
        \labeldescend{pastelGreen!40}{2,1}
        \labeldescend{pastelGreen!50}{2,0}
        \labelclimb{pastelGreen!40}{2,3}
        \labelclimb{pastelGreen!50}{2,4}
        \threesplitcolor{1,2}{pastelGreen!30}{pastelGreen!10}{pastelGreen!30}
        \threesplitcolor{1,3}{pastelGreen!50}{pastelGreen!30}{pastelGreen!10}
        \threesplitcolor{1,4}{pastelGreen!70}{pastelGreen!50}{pastelGreen!20}
        \threesplitcolor{1,1}{pastelGreen!10}{pastelGreen!30}{pastelGreen!50}
        \threesplitcolor{1,0}{pastelGreen!20}{pastelGreen!50}{pastelGreen!70}
        \fill[white, draw=black, line width=0.5pt] (-0.75, 4.25) rectangle (-0.25, 4.75);
        \node at (-0.5, 4.5) {\footnotesize $5$};
    \end{tikzpicture}
    };

    \node at (10,-6) {
    \begin{tikzpicture}[x=0.9cm,y=0.9cm]
        \casgrid
        \colorit{pastelRed}{3,2}
        \colorit{pastelGreen!50}{3,1}
        \colorit{pastelGreen}{3,0}
        \colorit{pastelGreen!50}{3,3}
        \colorit{pastelGreen}{3,4}
        \labeldescend{pastelGreen!30}{2,2}
        \labeldescend{pastelGreen!40}{2,1}
        \labeldescend{pastelGreen!50}{2,0}
        \labelclimb{pastelGreen!40}{2,3}
        \labelclimb{pastelGreen!50}{2,4}
        \labeldescend{pastelGreen!70}{1,0}
        \labeldescend{pastelGreen!50}{1,1}
        \labeldescend{pastelGreen!30}{1,2}
        \labelclimb{pastelGreen!50}{1,3}
        \labelclimb{pastelGreen!70}{1,4}
        \labeldescend{pastelGreen!90}{0,0}
        \labeldescend{pastelGreen!70}{0,1}
        \labeldescend{pastelGreen!50}{0,2}
        \labelclimb{pastelGreen!70}{0,3}
        \labelclimb{pastelGreen!90}{0,4}
        \fill[white, draw=black, line width=0.5pt] (-0.75, 4.25) rectangle (-0.25, 4.75);
        \node at (-0.5, 4.5) {\footnotesize $6$};
    \end{tikzpicture}
    };
\end{tikzpicture}
\end{center}
For each state-action pair, the transition model predicts the distribution over next states. Step 1 illustrates this process when taking the climb action in the state next to the collision state with thicker arrows representing higher probability transitions. In step 2, the expected value of taking each action is computed as a weighted average of the values of the next states. In step 3, the value of the current state is computed as the maximum expected value over all actions and the policy for the current state is updated to select the highest value action. Step 4 repeats this process for the remaining states in the current column. Steps 5 and 6 continue this process for the remaining columns until the values and policies for all states are computed. During this process, we maintain the value computed for each state-action pair and store these values in the logic table.
\end{textbox}

Given the states, actions, transition model, and reward function for an MDP, the goal is to find a policy that maximizes the expected cumulative reward over time. A \textit{policy} for an MDP maps states to actions. ACAS X solves for the optimal policy using dynamic programming \cite{Bellman}. First the state space is discretized into a finite set of states. As shown in the example, the dynamic programming solution works backwards in time starting from a time to collision of zero, computing the expected cumulative reward for each state-action pair. The optimal policy is then determined by selecting the action that maximizes the expected cumulative reward for each state. The result of the dynamic programming process is a logic table that stores the expected cumulative reward for each state-action pair.

\begin{figure}[t]
    \centering
    \includegraphics{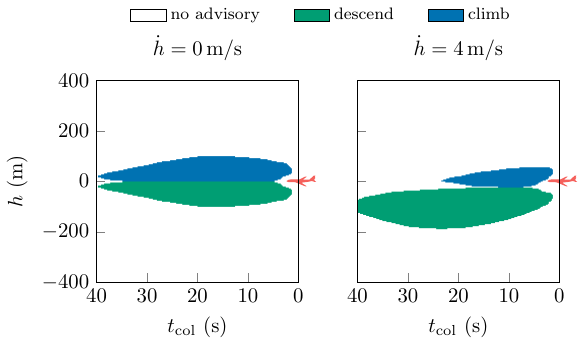}
    \caption{\label{fig:cas_policy_slices} Slices of the optimal policy for a notional example with three possible actions. In the plot on the left, both aircraft are in level flight, while in the plot on the right, the ownship is climbing. The previous action is fixed at no advisory in both plots. The colors represent the optimal action at each state, and the red aircraft represents the location of the intruder aircraft.}
\end{figure}

\Cref{fig:cas_policy_slices} shows slices of the optimal policy for a notional example with only three actions: climb, descend, and no advisory. The policy slices highlight multiple interesting features of the optimal policy. For example, the notch on the right side of both plots indicates a region where the optimal action is no advisory even though a collision is imminent. In this region, neither a climb or descend action is enough to avoid a collision, so the logic decides to take no action. When the aircraft are in level flight, a notch is also present on the left side of the optimal policy. Due to the uncertainty in the aircraft trajectories, it is better to wait to alert when the intruder is nearly coaltitude until it is clear whether the intruder will end up above or below the ownship.

The need to discretize the state space emphasizes the importance of selecting a compact representation for the state variables. It is important to ensure that the discretization is fine enough to capture the important features of the state space but coarse enough to ensure that the dynamic programming solution remains computationally tractable. Furthermore, the size of the final logic table that must be stored onboard the aircraft is determined by the number of total discrete state-action pairs. It is possible to compress the final logic table using standard compression techniques \cite{Kochenderfer2013a}, and recent work has explored the use of neural networks to further compress the logic table \cite{Julian2019jgcd,Julian2016dasc}. Other work has proposed solution techniques such as deep reinforcement learning that avoid the need to discretize the state space \cite{cone2023reward}. However, these techniques are often brittle and require careful tuning to ensure that the resulting logic is safe.

\subsubsection{Table Lookup Process}
To select an appropriate collision avoidance maneuver using the optimized logic table, ACAS X performs multilinear interpolation on the action values stored in the table at its current state and selects the action with the highest value. However, due to imperfect sensing, the exact state is unknown during operation. Instead, the surveillance module outputs a belief state that represents a probability distribution over possible states. The presence of this state uncertainty extends the MDP formulation to a partially observable Markov decision process (POMDP) formulation.
\begin{marginnote}
\entry{POMDP}{partially observable Markov decision process}
\end{marginnote}

To account for this uncertainty, ACAS X uses a POMDP solution technique called QMDP \cite{Chryssanthacopoulos2011jgcd}. QMDP solves the underlying MDP offline and uses the belief state online during the action selection process \cite{littman1995learning}. Using this technique, the table lookup process involves two steps. The first step is to use the belief state to compute a weighted average of the action values in the logic table. ACAS X then selects the action with the highest value. \Cref{fig:acasx_lookup} illustrates this process. The table lookup process requires minimal computation and can be performed efficiently in real time.

\begin{figure}
    \begin{center}
    \begin{tikzpicture}
        \node at (0, 0) {
        \begin{tikzpicture}
            \casgrid
            \threesplitcolor{1,2}{pastelGreen!30}{pastelGreen!10}{pastelGreen!30}
            \threesplitcolor{1,3}{pastelGreen!50}{pastelGreen!30}{pastelGreen!10}
            \threesplitcolor{1,4}{pastelGreen!70}{pastelGreen!50}{pastelGreen!20}
            \threesplitcolor{1,1}{pastelGreen!10}{pastelGreen!30}{pastelGreen!50}
            \threesplitcolor{1,0}{pastelGreen!20}{pastelGreen!50}{pastelGreen!70}

            \threesplitcolor{0,2}{pastelGreen!50}{pastelGreen!30}{pastelGreen!50}
            \threesplitcolor{0,3}{pastelGreen!70}{pastelGreen!50}{pastelGreen!30}
            \threesplitcolor{0,4}{pastelGreen!90}{pastelGreen!70}{pastelGreen!50}
            \threesplitcolor{0,1}{pastelGreen!30}{pastelGreen!50}{pastelGreen!70}
            \threesplitcolor{0,0}{pastelGreen!50}{pastelGreen!70}{pastelGreen!90}

            \threesplitcolor{2,2}{pastelGreen!30}{pastelGreen!20}{pastelGreen!30}
            \threesplitcolor{2,3}{pastelGreen!40}{pastelGreen!30}{pastelGreen!20}
            \threesplitcolor{2,4}{pastelGreen!50}{pastelGreen!40}{pastelGreen!30}
            \threesplitcolor{2,1}{pastelGreen!20}{pastelGreen!30}{pastelGreen!40}
            \threesplitcolor{2,0}{pastelGreen!30}{pastelGreen!40}{pastelGreen!50}

            \threesplitcolor{3,2}{pastelRed}{pastelRed}{pastelRed}
            \threesplitcolor{3,3}{pastelGreen!50}{pastelGreen!50}{pastelGreen!50}
            \threesplitcolor{3,4}{pastelGreen}{pastelGreen}{pastelGreen}
            \threesplitcolor{3,1}{pastelGreen!50}{pastelGreen!50}{pastelGreen!50}
            \threesplitcolor{3,0}{pastelGreen}{pastelGreen}{pastelGreen}

            \node at (1.5, 5) {Table};
                    
        \end{tikzpicture}
        };

        \node at (6, 0) {
        \begin{tikzpicture}
            \casgrid
            \threesplitcolor{1,2}{pastelGreen!30}{pastelGreen!10}{pastelGreen!30}
            \threesplitcolor{1,3}{pastelGreen!50}{pastelGreen!30}{pastelGreen!10}
            \threesplitcolor{1,4}{pastelGreen!70}{pastelGreen!50}{pastelGreen!20}
            \threesplitcolor{1,1}{pastelGreen!10}{pastelGreen!30}{pastelGreen!50}
            \threesplitcolor{1,0}{pastelGreen!20}{pastelGreen!50}{pastelGreen!70}

            \coloritopacb{white}{1,0}{0.6}{0.8}
            \coloritopacb{white}{1,1}{0.45}{1.1}
            \coloritopacb{white}{1,2}{0.2}{1.6}
            \coloritopacb{white}{1,3}{0.0}{2.0}
            \coloritopacb{white}{1,4}{0.2}{1.6}

            \threesplitcolor{0,2}{pastelGreen!50}{pastelGreen!30}{pastelGreen!50}
            \threesplitcolor{0,3}{pastelGreen!70}{pastelGreen!50}{pastelGreen!30}
            \threesplitcolor{0,4}{pastelGreen!90}{pastelGreen!70}{pastelGreen!50}
            \threesplitcolor{0,1}{pastelGreen!30}{pastelGreen!50}{pastelGreen!70}
            \threesplitcolor{0,0}{pastelGreen!50}{pastelGreen!70}{pastelGreen!90}

            \coloritopacb{white}{0,0}{0.75}{0.5}
            \coloritopacb{white}{0,1}{0.5}{1.0}
            \coloritopacb{white}{0,2}{0.3}{1.4}
            \coloritopacb{white}{0,3}{0.2}{1.6}
            \coloritopacb{white}{0,4}{0.3}{1.4}

            \threesplitcolor{2,2}{pastelGreen!30}{pastelGreen!20}{pastelGreen!30}
            \threesplitcolor{2,3}{pastelGreen!40}{pastelGreen!30}{pastelGreen!20}
            \threesplitcolor{2,4}{pastelGreen!50}{pastelGreen!40}{pastelGreen!30}
            \threesplitcolor{2,1}{pastelGreen!20}{pastelGreen!30}{pastelGreen!40}
            \threesplitcolor{2,0}{pastelGreen!30}{pastelGreen!40}{pastelGreen!50}

            \coloritopacb{white}{2,0}{0.75}{0.5}
            \coloritopacb{white}{2,1}{0.5}{1.0}
            \coloritopacb{white}{2,2}{0.3}{1.4}
            \coloritopacb{white}{2,3}{0.2}{1.6}
            \coloritopacb{white}{2,4}{0.3}{1.4}

            \threesplitcolor{3,2}{pastelRed}{pastelRed}{pastelRed}
            \threesplitcolor{3,3}{pastelGreen!50}{pastelGreen!50}{pastelGreen!50}
            \threesplitcolor{3,4}{pastelGreen}{pastelGreen}{pastelGreen}
            \threesplitcolor{3,1}{pastelGreen!50}{pastelGreen!50}{pastelGreen!50}
            \threesplitcolor{3,0}{pastelGreen}{pastelGreen}{pastelGreen}

            \coloritopacb{white}{3,0}{0.9}{0.2}
            \coloritopacb{white}{3,1}{0.6}{0.8}
            \coloritopacb{white}{3,2}{0.5}{1.0}
            \coloritopacb{white}{3,3}{0.45}{1.1}
            \coloritopacb{white}{3,4}{0.5}{1.0}

            \node at (1.5, 5) {Belief State};

            \draw[gray, line width=0.5] (1,0) -- (5,2);
            \draw[gray, line width=0.5] (1,1) -- (5,2);
            \draw[gray, line width=0.5] (1,2) -- (5,2);
            \draw[gray, line width=0.5] (1,3) -- (5,2);
            \draw[gray, line width=0.5] (1,4) -- (5,2);

            \draw[gray, line width=0.5] (0,0) -- (5,2);
            \draw[gray, line width=0.5] (0,1) -- (5,2);
            \draw[gray, line width=0.5] (0,2) -- (5,2);
            \draw[gray, line width=0.5] (0,3) -- (5,2);
            \draw[gray, line width=0.5] (0,4) -- (5,2);

            \draw[gray, line width=0.5] (2,0) -- (5,2);
            \draw[gray, line width=0.5] (2,1) -- (5,2);
            \draw[gray, line width=0.5] (2,2) -- (5,2);
            \draw[gray, line width=0.5] (2,3) -- (5,2);
            \draw[gray, line width=0.5] (2,4) -- (5,2);

            \draw[gray, line width=0.5] (3,0) -- (5,2);
            \draw[gray, line width=0.5] (3,1) -- (5,2);
            \draw[gray, line width=0.5] (3,2) -- (5,2);
            \draw[gray, line width=0.5] (3,3) -- (5,2);
            \draw[gray, line width=0.5] (3,4) -- (5,2);

            \threesplitcolorbig{5,2}{pastelGreen!55}{pastelGreen!35}{pastelGreen!10}

            \draw[stealth-, fignumcolor, line width=1.5] (5.6,2.4) -- (6.15,2.4);
                    
        \end{tikzpicture}
        };
        
        % \node at (9, -0.25) {
        % \begin{tikzpicture}[x=2cm, y=2cm]
        %     \threesplitcolorbig{0,0}{pastelGreen!55}{pastelGreen!35}{pastelGreen!10}
        %     % Highlight the chosen climb action with bold border
        %     \draw[fignumcolor, line width=4pt, rounded corners=2] 
        %         ($(0,0) - (0.35,-0.13333)$) rectangle ($(0,0) + (0.35,0.4)$);
        % \end{tikzpicture}
        % };
    \end{tikzpicture}
    \end{center}
    \caption{\label{fig:acasx_lookup} Notional example of the ACAS X table lookup process. States with thicker borders and low opacity in the belief state represent states with higher probability. The action values in the table are weighted by the belief state to compute the expected action values (right). The action with the highest expected value, indicated by the blue arrow, is selected.}
\end{figure}

To incorporate other system requirements that are difficult to encode in the MDP formulation, ACAS X also applies online costs to the interpolated action values during the table lookup process \cite{owen2019acas}. These costs encourage  behaviors such as inhibiting descend advisories during low-altitude operations and selecting maneuvers that are platform dependent. These costs may also be used for coordination of advisories between multiple aircraft, which we discuss in the next section.

\subsection{Coordination}\label{sec:coordination}
If both the ownship and intruder aircraft in an encounter are equipped with a collision avoidance system, it is beneficial to coordinate their actions to avoid potential conflicts. This coordination can take multiple forms. One option is to use a global coordination scheme, in which the aircraft share all sensor measurements and select the optimal joint action \cite{ikeda2002automatic,christodoulou2006automatic}. However, this coordination scheme poses challenges in situations with high uncertainty. Another option is to adopt a leader-follower paradigm, in which one aircraft selects its action first and the other aircraft selects a compatible action. While the leader-follower paradigm is suboptimal compared to global coordination, it is simpler to implement, less computationally expensive, and requires less communication.

TCAS and ACAS X both adopt the same leader-follower paradigm to ensure interoperability between existing systems \cite{Asmar2013atc408}, which is critical because TCAS will continue to exist well after the introduction of ACAS X. The aircraft with the lower Mode S address is designated the leader and selects its action first. The leader then sends a message to the follower that says either ``Do Not Climb'' or ``Do Not Descend'' for vertical logics and ``Do Not Turn Left'' or ``Do Not Turn Right'' for horizontal logics. The follower may then select any compatible action. For example, if the leader sends a ``Do Not Climb'' message, the follower may select any action corresponding to a down sense. ACAS X enforces these messages by applying large online costs to action values that correspond to incompatible actions.

For vertical logics, selecting which actions are compatible is straightforward. The leader and follower should always select actions of opposite sense. If the leader selects an action with a down sense, we always want the follower to select an action with an up sense. However, for horizontal logics, the situation is less clear. If the ownship and intruder are in an overtake pattern, we want them to turn opposite directions. If the ownship and intruder are in a head-on geometry, we want them to turn in the same direction. If they are in a crossing pattern, the optimal coordination message is less clear (\cref{fig:horiz_coord}). 

\begin{figure}[t] % horiz coord
    \begin{tikzpicture}
        \node at (0, 0) {
            \begin{tikzpicture}
                \draw[rounded corners=10, -stealth] (1.5,0) -- (1.5,1) -- (1.9, 1.5);
                \draw[rounded corners=10, -stealth, densely dashed, fignumcolor, line width=1] (0,0) -- (0,1) -- (-0.4, 1.5);

                \node [aircraft top,fill=black,draw=black,minimum width=1cm,rotate=90,scale=0.6] at (0,0) {};
                \node at (0, -0.6) {\footnotesize follower};

                \node [aircraft top,fill=black,draw=black,minimum width=1cm,rotate=90,scale=0.6] at (1.5,0) {};
                \node at (1.5, -0.6) {\footnotesize leader};

                \draw [rounded corners] (-1, -1) rectangle (2.5, 3.25);

                \node[fignumcolor] at (0.75, 2.75) {\textbf{Do Not Turn Right}};

            \end{tikzpicture}
        };

        \node at (4, 0) {
            \begin{tikzpicture}
                \draw[rounded corners=10, -stealth] (1.5,0) -- (1.5,1) -- (1.9, 1.5);
                \draw[rounded corners=10, -stealth, densely dashed, fignumcolor, line width=1] (0,1.5) -- (0,0.5) -- (-0.4, 0);

                \node [aircraft top,fill=black,draw=black,minimum width=1cm,rotate=270,scale=0.6] at (0,1.5) {};
                \node at (0, 2.1) {\footnotesize follower};

                \node [aircraft top,fill=black,draw=black,minimum width=1cm,rotate=90,scale=0.6] at (1.5,0) {};
                \node at (1.5, -0.6) {\footnotesize leader};

                \draw [rounded corners] (-1, -1) rectangle (2.5, 3.25);

                \node[fignumcolor] at (0.75, 2.75) {\textbf{Do Not Turn Left}};

            \end{tikzpicture}
        };

        \node at (8, 0) {
            \begin{tikzpicture}
                \draw[rounded corners=10, -stealth] (1.5,0) -- (1.5,1) -- (1.9, 1.5);

                \node [aircraft top,fill=black,draw=black,minimum width=1cm,rotate=0,scale=0.6] at (0,1.5) {};
                \node at (0, 2.0) {\footnotesize follower};

                \node [aircraft top,fill=black,draw=black,minimum width=1cm,rotate=90,scale=0.6] at (1.5,0) {};
                \node at (1.5, -0.6) {\footnotesize leader};

                \draw [rounded corners] (-1, -1) rectangle (2.5, 3.25);

                \node[fignumcolor] at (0.75, 2.75) {\textbf{?}};

            \end{tikzpicture}
        };
    \end{tikzpicture}
    \vspace{2mm}
    \caption{\label{fig:horiz_coord} Horizontal coordination scenarios. In an overtake scenario (left), the aircraft should turn in opposite directions. In a head-on scenario (middle), the aircraft should turn in the same direction. In a crossing scenario (right), the optimal coordination message is less clear.}
\end{figure}

The horizontal logic for ACAS Xu uses a separate coordination table that takes into account the relative positions and velocities of the aircraft to determine the optimal coordination message \cite{Tompa2016,Tompa2018}. The table is generated by formulating an MDP similar to the MDP used for optimizing the logic table. However, instead of RAs as the MDP actions, the actions are the possible coordination messages. During operation, the leader first looks up its RA in the horizontal logic table and then looks up its coordination message in the coordination table.

% Multiple possible schemes
    % Global coordination actions are selected jointly \cite{ikeda2002automatic} \cite{christodoulou2006automatic}
    % Leader-follower paradigm where leader selects action first and then follower selects compatible action

% How the coordination procedure is performed
    % Leader-follower paradigm where leader selects action first and then follower selects compatible action
% What is a compatible action?
    % Up sense and down sense
    % Horizontal less obvious

\subsection{Multithreat Scenarios}
Most airspace encounters involve only one intruder. However, on rare occasions, there may be multiple intruders present at the same time. In these multithreat scenarios, the collision avoidance system must select an action that accounts for all intruders. %Similar to the coordination strategies discussed in \cref{sec:coordination}, there are tradeoffs between global and local strategies \cite{kuchar2002review}. Global multithreat strategies consider all intruders at once, while local strategies focus consider pairwise interactions. 

%While global strategies are optimal, they are often computationally intractable. For this reason, TCAS and ACAS X both use local strategies that consider pairwise interactions. 

TCAS uses a command arbitration method that considers the RAs for each intruder and applies a set of heuristic rules to select a final RA. First, TCAS determines the optimal action for each intruder in isolation. If only one intruder results in an RA, TCAS issues that RA. If multiple intruders result in RAs of the same sense, TCAS issues the strongest selected RA. If multiple intruders result in RAs of different senses, TCAS applies a set of heuristic rules to select a final RA.

ACAS X uses a decomposition method known as utility fusion~\cite{Chryssanthacopoulos2012decomp}. In this process, ACAS X first obtains the expected utilities for each isolated pairwise interaction. The system then selects an action by fusing these utilities using a max-min strategy. For each possible action, ACAS X computes the minimum expected utility over all intruders. The action with the maximum of these minimum expected utilities is then selected. As shown in \cref{fig:multithreat}, this process encourages the logic to prioritize actions that result in high expected utility for all intruders.

\begin{figure}[t] % multithreat
    \begin{center}
    \begin{tikzpicture}
        \node at (0, 0) {
            \begin{tikzpicture}
                \draw[rounded corners] (-1.5, -1.75) rectangle (3.7, 1.75);
                \draw[-stealth, fignumcolor, line width=1] (0,0) -- (3,0);
                \node [aircraft top,fill=black,draw=black,minimum width=1cm,rotate=0,scale=0.6] at (0,0) {};

                \node [aircraft top,fill=gray,draw=gray,minimum width=1cm,rotate=180,scale=0.6] at (3,1) {};
                \node [aircraft top,fill=gray,draw=gray,minimum width=1cm,rotate=180,scale=0.6] at (3,-1) {};

                \threesplitcolorh{2.2, 1}{pastelRed!20}{pastelGreen!10}{pastelGreen!50}
                \threesplitcolorh{2.2, -1}{pastelGreen!50}{pastelGreen!10}{pastelRed!20}
                \threesplitcolorh{-0.8, 0}{pastelRed!20}{pastelGreen!10}{pastelRed!20}
                % \draw[fignumcolor, line width=1] (0.45,-0.13333) rectangle (1.15,0.13333);
                \draw[fignumcolor, line width=1] (-1.15,-0.13333) rectangle (-0.45,0.13333);

            \end{tikzpicture}
        };

        \node at (6.5, 0) {
            \begin{tikzpicture}
                \draw[rounded corners] (-1.5, -1.75) rectangle (5.7, 1.75);
                \draw[-stealth, rounded corners=20, fignumcolor, line width=1] (0,0) -- (1,0) -- (2,1);
                \node [aircraft top,fill=black,draw=black,minimum width=1cm,rotate=0,scale=0.6] at (0,0) {};

                \node [aircraft top,fill=gray,draw=gray,minimum width=1cm,rotate=180,scale=0.6] at (5.1,1) {};
                \node [aircraft top,fill=gray,draw=gray,minimum width=1cm,rotate=180,scale=0.6] at (3,-1) {};

                \threesplitcolorh{4.3, 1}{pastelGreen!20}{pastelGreen!30}{pastelGreen!50}
                \threesplitcolorh{2.2, -1}{pastelGreen!50}{pastelGreen!10}{pastelRed!20}
                \threesplitcolorh{-0.8, 0}{pastelGreen!20}{pastelGreen!10}{pastelRed!20}
                % \draw[fignumcolor, line width=1] (0.45,-0.13333) rectangle (1.15,0.13333);
                \draw[fignumcolor, rounded corners=2, line width=1] (-1.15,0.13333) rectangle (-0.45,0.4);

            \end{tikzpicture}
        };
    \end{tikzpicture}
    \end{center}
    \caption{\label{fig:multithreat} Notional examples of two multithreat scenarios. In the example on the left, the intruders (gray) are equidistant on opposite sides of the ownship (black). The isolated costs are shown next to each intruder, and the min-max fused costs are shown behind the ownship. In this case, turning left of right could have equally bad consequences, so the logic selects no manuever. In the example on the right, the intruder to the left of the ownship is further than the intruder to the right, and the logic decides to turn left.}
\end{figure}

% So far, have discussed just a single intruder
% Although rare, sometimes might be multiple intruders (we call this multithreat scenarios)
% Similar to coordination, could take global approach but computationally expensive \cite kuchar
% Instead, use pairwise methods that make use of single threat logic solution

% TCAS uses heuristic rules
% Look up actions for each intruder
    % If only one intruder has alert, do that
    % If all alerts in same sense, pick the strongest one
    % If alerts in different senses, use heuristic rules

% ACAS X uses utility fusion

% \section{OPERATIONAL SUITABILITY}\label{sec:operational_suitability}

\section{VALIDATION}\label{sec:validation}
A significant portion of the development effort for collision avoidance systems has focused on validating their performance to ensure safe deployment. As we discuss in \cref{sec:regulatory_acceptance}, validation is a critical step on the path to regulatory acceptance. Collision avoidance systems are typically validated using a variety of techniques that make use of simulation and real-world flight tests. Each technique has its own strengths and weaknesses. For example, one benefit of simulation-based testing is that it allows for the exploration of a wide range of scenarios and conditions that may be difficult, expensive, and unsafe to replicate in real-world tests. However, simulation-based testing may not capture the full complexity of real-world operations, so flight tests are also essential validation tools. This section begins with a focus on tools for simulation-based validation and concludes by describing techniques used for real-world flight testing.
%In this section, we discuss a variety of simulation-based and real-world flight test methodologies.

\subsection{Encounter Modeling}
Before applying simulation-based safety validation algorithms to a system, we must create a computational model of its environment \cite{Kochenderfer2026validation}. It is important to ensure that this model is accurate so that the validation results it produces are reliable. One key challenge in modeling the environment of airborne collision avoidance systems is capturing the dynamic behavior of other aircraft and their interactions with the ownship into an \textit{encounter model}. An encounter model is a probabilistic model that describes how aircraft behave in close proximity with one another. System designers use these models to generate pairwise encounters and test the behavior of their collision avoidance system in many scenarios.
\begin{marginnote}
\entry{encounter model}{probabilistic model that describes how aircraft behave in close proximity with one another}
\end{marginnote}

Airspace data is a critical component in the development of accurate encounter models. Encounter models developed to test TCAS starting in the 1980s were primarily based on radar data, which provided detailed information about aircraft trajectories \cite{mclaughlin1992safety,mclaughlin1997safety,MiquelRigottiEuroEncounter2002,chabert2005safety}. Before the development of ACAS X, these models were updated and extended using a principled statistical framework \cite{Kochenderfer2008em,Kochenderfer2008icns,Kochenderfer2008lljournal,Kochenderfer2010jgcd}. The data used to create the new models consisted of nine months of radar tracks that covered much of the continental United States.

The first step in producing an encounter model from radar data is to find encounters in the raw radar tracks and extract relevant variables. To create a probabilistic model, the next step is to fit a statistical distribution to the extracted variables, capturing the uncertainty and variability in aircraft behavior. It is important to ensure that we model the full joint distribution over the relevant variables by capturing the dependencies between them. For example, aircraft flying at higher altitudes tend to fly faster, and we want to ensure that the resulting model reflects these types of relationships. Bayesian networks provide a powerful framework for these types of representations and come with a variety of established structure learning algorithms that can uncover statistical dependencies between variables given a data set \cite{Kochenderfer2022}.

Airspace encounter models often consist of two Bayesian networks \cite{Kochenderfer2008em,Kochenderfer2008icns,Kochenderfer2008lljournal,Kochenderfer2010jgcd}. The first Bayesian network represents the distribution over initial states for the ownship and intruder. The second Bayesian network is a dynamic Bayesian network, which represents how variables at the current time step influence the variables at the next time step. The structure and parameters of each network can be learned from the processed radar data. To sample a new encounter, we first sample an initial state from the initial state network. Then, we use the dynamic Bayesian network to propagate the state forward in time, generating the encounter trajectories.

Encounter models have been customized to represent the behavior of a variety of aircraft types and airspace procedures. They can be correlated \cite{Kochenderfer2008cor} or uncorrelated \cite{Kochenderfer2008uncor,Weinert2013}. Correlated encounter models involve transponder-equipped aircraft that are in contact with ATC. Since ATC intervention often leads to correlation between the aircraft trajectories, these models capture this correlation by generating both the ownship and intruder trajectories in a joint manner. In constrast, uncorrelated encounter models represent aircraft following visual flight rules with no coordinated intervention. Therefore, trajectories from these models are be sampled independently and combined to create an encounter.

Uncorrelated encounter models have been created for a variety of aircraft types including small unmanned aircraft systems (UAS) \cite{weinert2020representative}, hobbyist unmanned aircraft \cite{Mueller2016comp}, urban air mobility aircraft \cite{Katz2019}, helicopter air ambulances \cite{weinert2018well}, and unconventional aircraft such as ultralights, gliders, balloons, and airships \cite{Edwards2009}. The data for these models comprises a variety of sources from radar data \cite{Edwards2009} to infrastructure maps \cite{weinert2020representative}. As new aircraft types are introduced into the airspace, we may have insufficient data to create accurate encounter models. In these cases, we must rely on expert knowledge \cite{Katz2019}. It is important that we continue to expand and update this models in the future as the airspace continues to evolve. Many of these encounter models have been made publicly available at \href{https://github.com/Airspace-Encounter-Models}{https://github.com/Airspace-Encounter-Models}.

% small UAS \cite{weinert2020representative}
% helicopter air ambulances \cite{Weinert2018well}
% unconventional aircraft such as ultralights, gliders, balloons, and airships \cite{Edwards2009}
% UAM \cite{Katz2019}

% Correlated vs uncorrelated
% Aircraft types
% Different data sources for other aircraft types (rotorcraft and small uas)
% Or no data (UAM)
% Also, some encounter sets may not represent the real-world statistically and may be designed, for example, to stress test
% Also might play back real world encounters

\subsection{Evaluation Metrics}
Collision avoidance systems are evaluated using several metrics related to safety, operational suitability, and pilot acceptance \cite{Holland2013}. One of the most common safety metrics is the probability of a near mid-air collision (NMAC). An NMAC is defined as an event in which two aircraft come within less than \SI{100}{\foot} vertically and \SI{500}{\foot} horizontally of one another. This definition was originally specified based on the typical shapes and sizes of manned aircraft; however, recent work has proposed smaller NMAC volumes for smaller UAS \cite{weinert2022near,alvarez2019acas}. Because determining whether an NMAC occurs does not require knowledge of the geometric shapes of each aircraft, using the number of NMACs as a proxy for the number of mid-air collisions (MACs) allows encounter simulators to model the aircraft as point masses. The probability of a MAC given that an NMAC occurs is estimated to be lower than \num{0.1} \cite{Kochenderfer2010atio}. The \textit{risk ratio} of a collision avoidance system is defined as the ratio of the probability of an NMAC when the system is in use to the probability of an NMAC when no system is in use. The risk ratio has been accepted internationally as a safety metric \cite{ICAOAnnex10VolIV2007}.
\begin{marginnote}
\entry{NMAC}{near mid-air collision}
\end{marginnote}

Beyond safety, operational suitability and pilot acceptance are important aspects of the effectiveness of collision avoidance systems. The primary operational suitability metric is the frequency of alerts in non-safety-critical encounters. It is important to minimize unnecessary alerts to avoid potential fatigue and distrust among pilots. Other metrics assess the frequency of disruptive alerts that require large changes in vertical rates, the frequency of reversals, and the frequency of alerts requiring the aircraft to cross altitudes before closest point of approach. Ensuring that reversals and crossings occur infrequently is critical for maintaining pilot trust.

% Interop?

\subsection{Computational Validation}

Given an airspace encounter model and a set of evaluation metrics, we can use simulation techniques to validate the performance of collision avoidance systems. One common approach is to use Monte Carlo simulation to generate a large number of encounters from the encounter model and evaluate the performance of the system in each encounter. \Cref{fig:example_encounter} shows an example of a simulated encounter. By aggregating the results, we can estimate metrics such as the probability of NMAC \cite{zeitlin2006collision}. One challenge with this approach is that NMACs are rare events, so a large number of encounters may be required to obtain accurate estimates. For example, if the probability of an NMAC is \num{e-9}, we must simulate billions of encounters to obtain a reliable estimate.

\begin{figure}
    \centering
    \input{vertical_logic_example.tex}
    \caption{\label{fig:example_encounter} Example of a simulated encounter between two aircraft. The aircraft on the left has a collision avoidance system and receives a series of RAs to avoid an NMAC.}
\end{figure}

To address the challenge of estimating the probability of rare failure events, we can use importance sampling techniques \cite{Kochenderfer2026validation}. Importance sampling techniques bias the encounter generation process towards encounters that are more likely to result in NMACs \cite{Chryssanthacopoulos2010}. We can then reweight the results to account for the bias introduced by the importance sampling and obtain an unbiased estimate of the probability of NMAC. However, selecting an appropriate importance sampling distribution can be difficult. The cross entropy method provides a systematic way to adapt the importance sampling distribution over multiple iterations \cite{Kim2016}.

In addition to estimating nominal failure probabilities, we can also use simulation techniques to understand the behavior of collision avoidance systems under a variety of challenging scenarios. These stress testing techniques can help identify potential weaknesses in the system and guide future improvements. A common stress testing technique is to vary encounter parameters such as vertical rates, acceleration rates, and miss distances to create worst-case situations for the collision avoidance system \cite{chludzinski2009evaluation}. However, it may be difficult to enumerate all possible challenging scenarios by hand. A technique called adaptive stress testing (AST) uses reinforcement learning to automatically identify the most likely failure scenarios \cite{Lee2020}. Finding potential failure modes for aircraft collision avoidance systems was one of the first use cases for AST \cite{Lee2018,Lee2018}, it has since been extended to other uses cases such as flight management systems \cite{Moss2020}, autonomous driving systems \cite{Corso2019itsc}, and large language models \cite{Hardy2025}.

Because failure probability estimation and stress testing techniques do not necessarily enumerate all possible scenarios, they do not produce guarantees about system performance. Formal methods can provide such guarantees by considering all possible outcomes under a set of assumptions. Several formal methods techniques have been applied to collision avoidance systems including hybrid systems modeling \cite{livadas2002high,jeannin2015formally}, game theoretic methods \cite{cleaveland2022formally}, and Petri net-based approaches \cite{jun2015analysis}. 

Furthermore, the neural network-based compression technique for the ACAS X logic tables \cite{Julian2019jgcd} became one of the first application areas for the field of neural network verification \cite{Julian2019dasc,Julian2019vnn,Irfan2020}. Neural network verification techniques use formal methods to provide properties of neural networks \cite{Liu2021}. For example, we can use a neural network verification tool to check that the neural network will always recommend an advisory with an up sense when the ownship and intruder are in level flight and the intruder is below the ownship. This prelimary work served as a catalyst for the field of neural network verification, which has since been applied to a variety of safety-critical systems \cite{Katz2021reluplex,brix2023first}.

% Given encounter model, can generate encounters in Monte Carlo fashion and validate performance
% Problem with rare failure events
% Use importance sampling where sample from different distribution
% May be difficult to hand pick so cross entropy

% Might also want to stress test the system in difficult scenarios
% TCAS (FTEG)
% AST
    % Use in other fields as well

% Failure probability estimation and stress testing do not necessarily enumerate all possible scenarios so do not necessarily produce guarantees
% Formal methods can produce guarantees under a set of assumptions
% Formal methods have been particularly interesting for the neural network compression
% Catalyst for the field of neural network verification

% Using encounter model to validate \cite{zeitlin2006collision}
% Importance sampling \cite{Chryssanthacopoulos2010}
% Cross entropy \cite{Kim2016}

% Formal methods \cite{jeannin2015formally}
% Stress testing
    % TCAS (FTEG) \cite{chludzinski2009evaluation}
    % AST \cite{Lee2018,Lee2018Diff}
% Colored Petri Net? \cite{jun2015analysis} - this is some sort of formal method - TCAS
% NNV \cite{manzanas2023evaluation}
% Formal verification \cite{jeannin2015formal} game theory \cite{cleaveland2022formally} TCAS \cite{livadas2002high}
% Neural network \cite{julian2019dasc,julian2019vnn,irfan2020}
    % in fact, has been catalyst for the field of nnv \cite{Katz2021reluplex,Liu2021,brix2023first}

% Sensitivity analysis \cite{Kochenderfer2010dasc}

% Cyber attacks? \cite{smith2022understanding}

% Continuous monitoring and adaptation
    % TCAS \cite{olson2010tcas}

\subsection{Flight Testing}
Beyond simulation-based testing, flight testing plays an important role in validating the safety and operational performance of a collision avoidance system. There are two major types of flight test activities typically associated with collision avoidance system development. The first activity is developmental flight testing, which involves testing the system in a controlled airspace environment by flying a set of scripted encounters. The second activity is operational flight testing, which involves testing the system in real-world operations.

Developmental flight testing is typically performed during the collision avoidance system development process. It serves many purposes such as concept validation, systems integration, initial pilot acceptability assessments, and validation of the modeling and simulation tools used for computational validation \cite{HollandACASX2013}. This type of testing provides engineering data with a goal of improving system performance; however, it is not intended to demonstrate that the system under test meets rigorous safety performance requirements. 

Due to cost, schedule, and safety considerations, developmental flight testing is limited to a small number of encounters and geometries in which safe separation can be maintained at all times. However, the results are critical to illuminate issues that require updates to the current system design and testing methodologies. For example, during ACAS Xa development, the team conducted two major developmental flight tests. They flew a total of 279 scripted encounters and used the data to update and validate the modeling and simulation tools, uncover important pilot feedback on alerting, and identify several performance issues to be addressed in the development process \cite{HollandACASX2013}.

Once a collision avoidance system reaches a state of relative maturity, operational test and evaluation is often conducted using line pilots. These pilots fly with the system over an extended period of time in non-scripted encounters in the actual operational environment. The primary goal of this flight test activity is to ensure that alerting is acceptable during actual operations across a broad range of conditions.  Operational test and evaluation was a key pre-deployment activity for both TCAS and ACAS Xa. The TCAS Limited Implementation Program \cite{HoneywellTCASLIP1989} identified several surveillance issues and resulted in a number of alerting changes, including raising the low altitude alerting limit. The ACAS Xa operational evaluation was conducted over \num{9} months, flying ACAS Xa across the United States. This evaluation revealed needed updates to the TA logic and the need for some additional adjustments to tracker algorithms \cite{TellerACASXA2017}.

\subsection{Validation and Adaptation Cycle}\label{subsec:cycle}
Rather than occurring once during the development process, validation and adaptation should be viewed as an ongoing cycle. For example, ACAS X required a \num{15}-month iterative tuning process that alternated between evaluating performance metrics using computational validation techniques and using the results to adapt the collision avoidance logic by hand or using an automated optimization process \cite{Holland2013}.
One challenge with this tuning approach is the expense of rerunning computational validation every time the logic is updated. 

To increase computational efficiency, surrogate modeling techniques can be employed to optimize objectives without the need for extensive reevaluation for each update \cite{smith2013collision}. However, it is important to ensure that the system is optimized for the proper tradeoff between safety and operational performance that meets the needs of stakeholders such as the Federal Aviation Administration (FAA),  the Single European Sky Air Traffic Management Research project, potential commercial vendors, and pilot associations. Preference elicitation techniques can be used to infer the stakeholders desired balance between safety and operational performance with minimal cognitive burden \cite{Lepird2015}. 

% Wes
Once the system has been deployed, system acceptance is facilitated by the identification and resolution of performance issues through system monitoring, issue reporting and analysis. 
%These activities work to ensure that system deficiencies or undesired behaviors are addressed by changes to the system, procedures, or training.  
System monitoring is enabled by the automatic reporting feature built into collision avoidance systems. By design, every time an RA is generated, TCAS and ACAS X broadcast information on the RA and the associated intruder aircraft. Research organizations, operators, and regulators collect and analyze this information to identify potential system deficiencies that warrant changes to the system, procedures, or training. For example, the TCAS Operational Performance Assessment program gathers these RA downlinks along with associated surveillance on nearby aircraft to understand TCAS alerting behavior, identify issues, and identify the impact of TCAS alerting on specific encounters \cite{olson2010tcas}.  

% continuous monitoring and adaptation should be implemented to ensure that it continues to meet safety and performance requirements in the evolving operational environment \cite{olson2010tcas}.

% For example, ACAS X required a 15 month iterative tuning process
    % Created next version, evaluated performance metrics using computational validation techniques, manually or automated process updated and repeat
% This can be expensive to rerun computational validation every time so surrogate modeling to optimize objectives
% But must know how to balance objectives properly when performing surrogate optimization
% During this process, it is important to keep in mind stakeholders (consisting of the FAA, the Single European Sky Air Traffic Management Research project, potential commercial vendors, and pilot associations.)
    % Preference elicitation! \cite{Lepird2015}
% During deployment, continue monitoring \cite{olson2010tcas}

\section{REGULATORY ACCEPTANCE}\label{sec:regulatory_acceptance}
Given the critical role of collision avoidance systems in maintaining aviation system safety, a structured regulatory process ensures that any new or modified collision avoidance system will meet required safety and performance objectives. Because aviation is a global, interconnected enterprise, these regulatory processes must be standardized and interoperable for all operations across the global airspace. Therefore, any new collision avoidance must be standardized within the international community, approved by worldwide aviation authorities, and accepted by pilots, controllers, and the flying public. 
%We discuss these processes in the remainder of this section.

\subsection{Standardization}
Collision avoidance systems are typically developed through a consensus-based standards development process that involves the full range of community stakeholders. These stakeholders include regulators such as the FAA, avionics and airframe manufacturers, pilot and air traffic controller unions, system designers, and aircraft operators. The collision avoidance standards are published in the form of Minimum Operational Performance Standards (MOPS) which describe the system goals, operational environment, required performance, and test performance. The MOPS also provide prescriptive logic in the form of pseudocode for TCAS and Julia code for ACAS X.
\begin{marginnote}
\entry{MOPS}{Minimum Operational Performance Standards}
\end{marginnote}

The RTCA (formerly known as the Radio Technical Commission for Aeronautics) in the United States and the European Organisation for Civil Aviation Equipment (EUROCAE) are the standards organizations that develop these MOPS. Typically, these MOPS are published jointly to ensure harmonization and international interoperability. During the CAS design process, the design team provides regular updates on the system design and associated safety and alerting performance to the RTCA and EUROCAE committees for discussion and consensus approval. These performance metrics are generated using the algorithms discussed in \cref{sec:validation} and fit into the validation cycle described in \cref{subsec:cycle}.

In addition to the design team, the collision avoidance system is often independently evaluated by others in the community and the results of these evaluations are also discussed by the committee \cite{Lebron1983}.  Once community consensus is reached, the resulting MOPS are published by RTCA and EUROCAE to serve as the basis for regulatory approval. The MOPS for TCAS are published as RTCA DO-185B \cite{RTCADO185B2013}, and the MOPS for ACAS Xa are published as RTCA DO-385A \cite{RTCADO3852019}.

\subsection{Certification}
Once the MOPS have been finalized, the collision avoidance system moves on to regulatory approval. The regulatory approval process ensures that the avionics, installation and integration on the aircraft, and the operational use of the collision avoidance system is safe and effective. This process is performed by civil aviation authorities such as the FAA and EASA and requires multiple steps.

The first step in this process is to provide instructions for manufacturers to implement the MOPS in their individual avionics units. These instructions are provided in the form of a Technical Standard Order (TSO) in the United States and a European Technical Standard Order (ETSO) in Europe. With some additions, deletions, or clarifications to the MOPS from RTCA and EUROCAE, the TSO and ETSO provide the necessary guidance for manufacturers to develop compliant systems. They ensure that the collision avoidance avionics meets the desired safety and operational performance. FAA TSO C-119e and TSO C-219 cover TCAS II v7.1 and ACAS Xa/Xo respectively.

In addition to the TSO and ETSO, regulatory agencies outline the required equipment for collision avoidance along with the associated airworthiness considerations and tests. These requirements are published in the form of an airworthiness advisory circular (AC). When manufacturers seek approval for their systems, they must demonstrate compliance with the requirements in the AC. If they comply with the airworthiness AC, they receive a type certificate (TC) or supplemental type certificate (STC), which grants airworthiness approval for installation on a specific aircraft type. The FAA AC 20-151C addresses TCAS v7.1 approvals.

Beyond software and equipment, regulatory guidance also covers pilot and controller training and procedures. Standardized and harmonized procedures are critical to ensure interoperability between pilots and controllers in the broader airspace system. For example, pilot procedures give priority to collision avoidance RAs over ATC instructions, and ATC procedures mandate controllers to cease giving separation guidance to aircraft once they are informed that an aircraft is responding to an RA. FAA guidance for operational procedures and training is published in FAA AC 120-55C, and controller procedures are contained in FAA Order 7110.65BB.

Outside of the FAA and EASA, the International Civil Aviation Organization (ICAO) contributes to international harmonization by providing a forum for international consensus and information sharing at regular technical meetings. They also publish Standards and Recommended Practices (SARPS), which member nations agree to follow unless they publish specific exemptions. ICAO guidance for the required functionality and associated performance of collision avoidance systems is published in ICAO Annex 10, Volume IV~\cite{ICAOAnnex10VolIV2018}. This guidance is aligned with RTCA and EUROCAE MOPs and their associated TSOs and ACs. Additionally, ICAO publishes pilot training and procedural guidance in ICAO Doc 8168 (PANS-OPS) \cite{ICAODoc8168VolIII2021} and controller guidance in ICAO Doc 4444 (PANS-ATM) \cite{ICAODoc98632021}.

\subsection{Societal and User Acceptance}
The standardization and certification processes are intended to ensure that the collision avoidance system will provide the desired level of safety and operational performance. In contrast to the processes involved in standardization and certification, societal and user acceptance is less formal. It is largely related to facilitating knowledge of system operation and ongoing monitoring to identify and resolve operational issues before they erode pilot and controller trust in the system.

Beyond the pilot and controller training prescribed by civil aviation authorities, additional information is published by regulators, research organizations and ICAO on collision avoidance operation. This information is published in the form of informational booklets that provide a succinct description of the system, expected safety and alerting performance, and pilot and controller procedures. The FAA has published a guide for TCAS v7.1 \cite{FAATCASII72011} and EUROCONTROL has published a similar guide for TCAS v7.1 and ACAS Xa/Xo \cite{EUROCONTROLACASGuide52025}. ICAO also has also published a manual in ICAO Doc 9863 \cite{ICAODoc98632021}.  These manuals are intended to help pilots, controllers, and the broader community understand the operation, limitations, and expected performance of the collision avoidance system.

As described in \cref{subsec:cycle} system monitoring is also an important aspect of societal and user acceptance. These activities work to ensure that system deficiencies or undesired behavior are addressed by changes to the system, procedures or training. This monitoring is performed by several agencies such as the FAA \cite{olson2010tcas}, EUROCONTROL \cite{EUROCONTROLACASXaCP1Validation2022}, aircraft operators, and avionics vendors. Taken together, these monitoring activities have identified a number of issues related to TCAS performance and allowed for an understanding of the interaction between TCAS alerting and airspace structure. These results in turn allowed the ACAS X developers to dramatically reduce undesired alerting caused by the incompatibility of TCAS alerting logic with the existing airspace structure.

\section{FUTURE DIRECTIONS}
Since their introduction, aircraft collision avoidance systems have significantly improved the safety of the airspace. However, there are still opportunities for further improvements as our airspace continues to evolve. The tragic collision between an American Eagle regional jet and a United States Army helicopter over Washington, D.C. in January 2025 ended a nine-year stretch without a jet transport mid-air collision occurring anywhere in the world. While TCAS performed as intended in this accident, it underscores the need for ACAS Xr, which will extend collision avoidance capability to rotorcraft.
% and underscores the need to pursue additional safety enhancements as traffic levels continue to rise. 

The mid-air collision risk for general aviation aircraft is significantly higher than that for the jet transport aircraft shown in \cref{fig:traffic_and_collisions}. In the United States, general aviation mid-air collisions have been occurring at the rate of approximately one in every \num{4} million flight hours, which is \num{160} times higher than what is shown in \cref{fig:traffic_and_collisions}. This increased rate is due to multiple factors including frequent reliance on visual flight procedures and see and avoid, high traffic densities near airports, and lack of widely-deployed collision avoidance technology in general aviation cockpits. 

Significant opportunities for safety improvement remain in the areas of airspace design, procedures, traffic situation awareness, and collision avoidance technology. One area of ongoing work is automatic return to flight path after a potential midair collision has been resolved. For manned aircraft, the pilots are responsible for returning the aircraft to its original flight path after responding to an RA. However, unmanned aircraft may require automated techniques that allow them to safely return to their original flight path without causing additional conflicts. Another area for further study is the integration of collision avoidance systems with other high-level decision-making systems onboard the aircraft. For example, current research is exploring the use of large language models to monitor safety and provide context-aware decision support \cite{Schlichting2025ecai}. 

The extensive research in aircraft collision avoidance spanning several decades makes a compelling case study for advanced decision-making systems operating in high-stakes settings. Many of the challenges involved in creating a robust collision avoidance system are present in a variety of other domains including maritime, ground vehicles, and spacecraft operations. For example, as the density of satellites and debris in orbit continues to increase, the National Aeronautics and Space Administration (NASA) is beginning to assess best practices for conjunction assessment and collision avoidance on orbit \cite{NASAConjunctionAssessmentHandbook2023}. The lessons learned from aircraft collision avoidance can inform the development of these systems.

Finally, the regulatory processes for collision avoidance systems provide a useful case study for the certification of advanced decision-making systems. The transition from the TCAS system with explicitly defined logic rules to ACAS X, which encodes the optimized logic in a lookup table of costs, represented a significant step forward in the regulatory acceptance of more complex decision-making systems. As autonomy continues to progress, systems are becoming increasingly complex and are starting to incorporate machine learning techniques such as neural networks. The certification of these systems will require new regulatory processes that consider the broader system context in which they operate \cite{durand2023formal}. The experience gained from the certification of collision avoidance systems can inspire these new processes.

\section*{ACKNOWLEDGMENTS}
We gratefully acknowledge Mr. Neal Suchy, FAA Collision Avoidance Program Manager, for his leadership, vision and dedication to aviation safety.  Mr. Suchy has funded development of all four ACAS X variants and has spearheaded the push for US and ICAO guidance to allow international harmonization and use of collision avoidance system. Without his efforts, ACAS X could not have moved from concept to an operational system.

\section*{DISTRIBUTION STATEMENT}
Approved for public release. Distribution is unlimited.
This material is based upon work supported by the Federal Aviation Administration under Air Force Contract No. FA8702-15-D-0001 or FA8702-25-D-B002. Any opinions, findings, conclusions or recommendations expressed in this material are those of the author(s) and do not necessarily reflect the views of the Federal Aviation Administration.

\printbibliography

\end{document}